\documentclass[10pt,journal,compsoc]{IEEEtran}
\usepackage{graphicx}
\usepackage{booktabs}
\usepackage{algorithm}
\usepackage{rotating}
\usepackage{balance}
\usepackage{color}
\usepackage{colortbl}

\usepackage{tabularx}
\usepackage{subfigure}
\usepackage{bm}
\usepackage{multirow}
\usepackage{enumitem}
\usepackage{amsmath}
\usepackage{amsfonts}
\usepackage{soul}
\usepackage{url} 

\usepackage{xcolor}

\ifCLASSOPTIONcompsoc
  \usepackage[nocompress]{cite}
\else
  \usepackage{cite}
\fi
\ifCLASSINFOpdf
\else
\fi
\begin{document}
%
\title{Memory-Guided Multi-View Multi-Domain\\ Fake News Detection}

\author{Yongchun~Zhu,
        Qiang~Sheng,
        Juan~Cao,
        Qiong~Nan,
        Kai~Shu,
        Minghui~Wu,
        Jindong~Wang,
        and~Fuzhen~Zhuang
\IEEEcompsocitemizethanks{\IEEEcompsocthanksitem Yongchun Zhu, Qiang Sheng, Juan Cao and Qiong Nan are with Key Lab of Intelligent Information Processing of Chinese Academy of Sciences (CAS), Institute of Computing Technology, CAS, Beijing 100190, China, and University of Chinese Academy of Sciences, Beijing 100049, China.
E-mail: \{zhuyongchun18s, shengqiang18z, caojuan, nanqiong19z\}@ict.ac.cn
\IEEEcompsocthanksitem Kai Shu is with Illinois Institute of Technology, Chicago, IL 60616, USA.
E-mail: kshu@iit.edu
\IEEEcompsocthanksitem Minghui Wu is with School of Computer and Computing Science, Zhejiang University City College, Hangzhou 310015, China.
E-mail: mhwu@zucc.edu.cn
\IEEEcompsocthanksitem Jindong Wang is with Microsoft Research Asia.
E-mail: jindong.wang@microsoft.com
\IEEEcompsocthanksitem Fuzhen Zhuang is with Institute of Artificial Intelligence, Beihang University, Beijing 100191, China, and SKLSDE, School of Computer Science, Beihang University, Beijing 100191, China.
E-mail: zhuangfuzhen@buaa.edu.cn
\IEEEcompsocthanksitem Corresponding Author: Juan Cao
}
}

%
%

\markboth{Journal of \LaTeX\ Class Files,~Vol.~14, No.~8, August~2015}%
{Shell \MakeLowercase{\textit{et al.}}: Bare Demo of IEEEtran.cls for Computer Society Journals}
%



\IEEEtitleabstractindextext{%
\begin{abstract}
The wide spread of fake news is increasingly threatening both individuals and society. Great efforts have been made for automatic fake news detection on a \textit{single} domain (e.g., politics). However, correlations exist commonly across multiple news domains, and thus it is promising to simultaneously detect fake news of \textit{multiple} domains. Based on our analysis, we pose two challenges in multi-domain fake news detection: 1) \textbf{\textit{domain shift}}, caused by the discrepancy among domains in terms of words, emotions, styles, etc. 2) \textbf{\textit{domain labeling incompleteness}}, stemming from the real-world categorization that only outputs one single domain label, regardless of topic diversity of a news piece. In this paper, we propose a Memory-guided Multi-view Multi-domain Fake News Detection Framework (M$^3$FEND) to address these two challenges. We model news pieces from a multi-view perspective, including semantics, emotion, and style. Specifically, we propose a Domain Memory Bank to enrich domain information which could discover potential domain labels based on seen news pieces and model domain characteristics. Then, with enriched domain information as input, a Domain Adapter could adaptively aggregate discriminative information from multiple views for news in various domains.
Extensive offline experiments on English and Chinese datasets demonstrate the effectiveness of M$^3$FEND, and online tests verify its superiority in practice. Our code is available at \url{https://github.com/ICTMCG/M3FEND}.
\end{abstract}

}

\maketitle

\IEEEdisplaynontitleabstractindextext

%
\IEEEpeerreviewmaketitle

\section{Introduction}
With the rapid growth of web technologies, more and more people rely on online social media for news acquisition. According to the 2021 Pew Research Center survey, 48\% of American adults get news from social media ``often'' or ``sometimes''~\cite{newsreport}. Meanwhile, the wide spread of fake news on social media has threatened both individuals and society. 
In the COVID-19 infodemic~\cite{naeem2020covid}, thousands of fake news pieces have been widely spread around the world~\cite{wikipedia}, which caused social panic~\cite{chinese_rumor} and weakened the effect of pandemic countermeasures~\cite{bursztyn2020misinformation}. Under such severe circumstances, automatic detection of fake news has been critical for the sustainable and healthy development of news platforms~\cite{shu2017fake}, and great efforts have been made to build automatic fake news detection systems~\cite{zhouxing2015,defend-system,sheng-acl22,zhu2022generalizing}. 

\begin{figure}[t]
\setlength{\abovecaptionskip}{3pt}
\setlength{\belowcaptionskip}{5pt}
	\centering
	\begin{minipage}[b]{1\linewidth}
		\centering
		\includegraphics[width=0.95\linewidth]{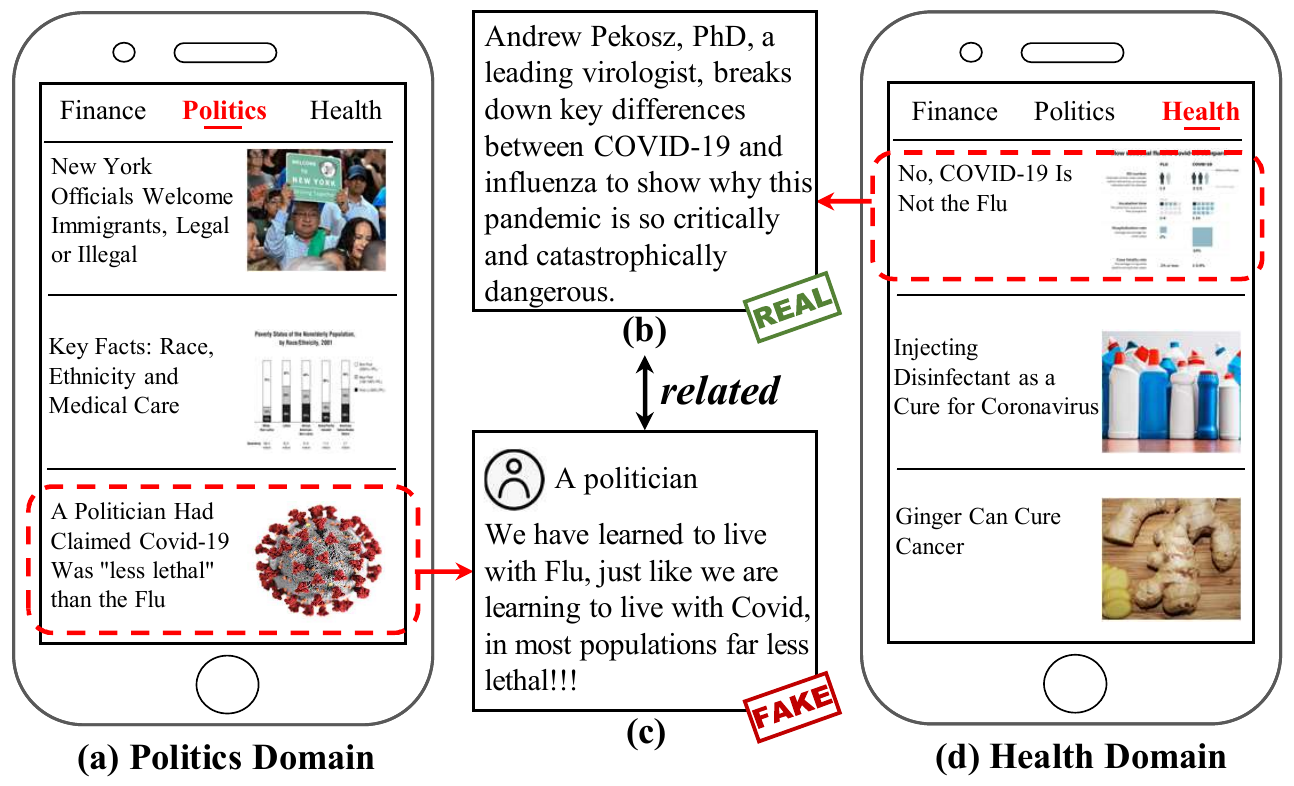}
	\end{minipage}
	\caption{An example of a real-world news platform with \textbf{\textit{multiple news domains}}. The news distributions vary from domain to domain, leading to the challenge of \textbf{\textit{domain shift}}. However, a news piece is a mixture of diverse elements which makes it relate to multiple news domains, e.g., the political news (c) is also related to the health news (b), leading to the challenge of \textbf{\textit{domain labeling incompleteness}}.}\label{fig:intro}
\end{figure}

A real-world news platform releases millions of news pieces in diverse domains every day, e,g., finance, politics, health, as shown in Figure~\ref{fig:intro}. However, most existing methods only focus on a \emph{single} domain (e.g. politics).
In fact, news pieces in different domains are inherently correlated, where news (b) and (c) come from different domains while sharing a similar topic of COVID-19.
Intuitively, simultaneously modeling multiple correlated news domains benefits fake news detection, and thus it is valuable to study multi-domain fake news detection in real-world news platforms~\cite{sheng2022characterizing}.

\begin{figure*}[t]
\begin{minipage}{.4\linewidth}
\subfigure[Top 20 words in the nine domains]{
\centering
\includegraphics[width=1.\columnwidth,height=.8\columnwidth]{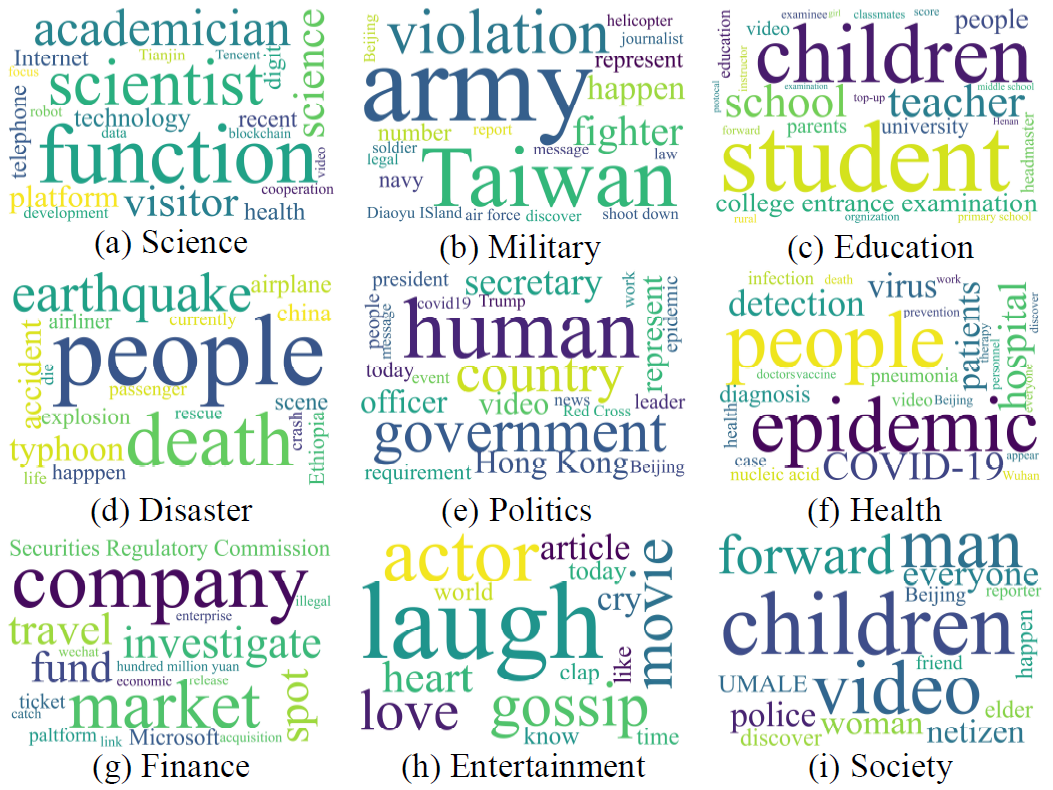}
\label{fig:word_usage}
}
\end{minipage}%
\begin{minipage}{.6\linewidth}
\subfigure[Style features of fake news]{
\centering
\includegraphics[width=0.5\columnwidth]{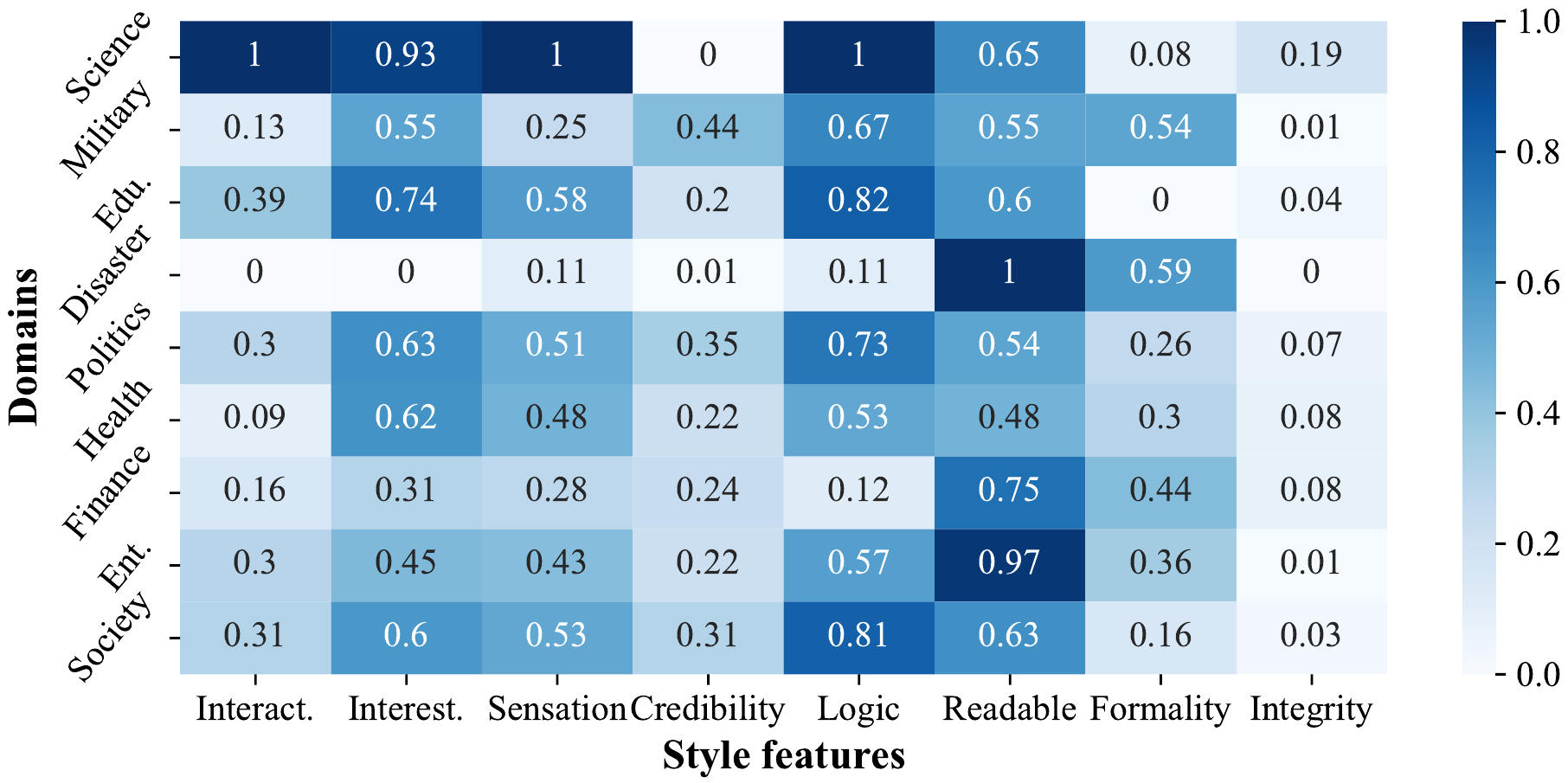}
\label{fig:analysis_a}
}
\subfigure[Emotion features of fake news]{
\centering
\includegraphics[width=0.5\columnwidth]{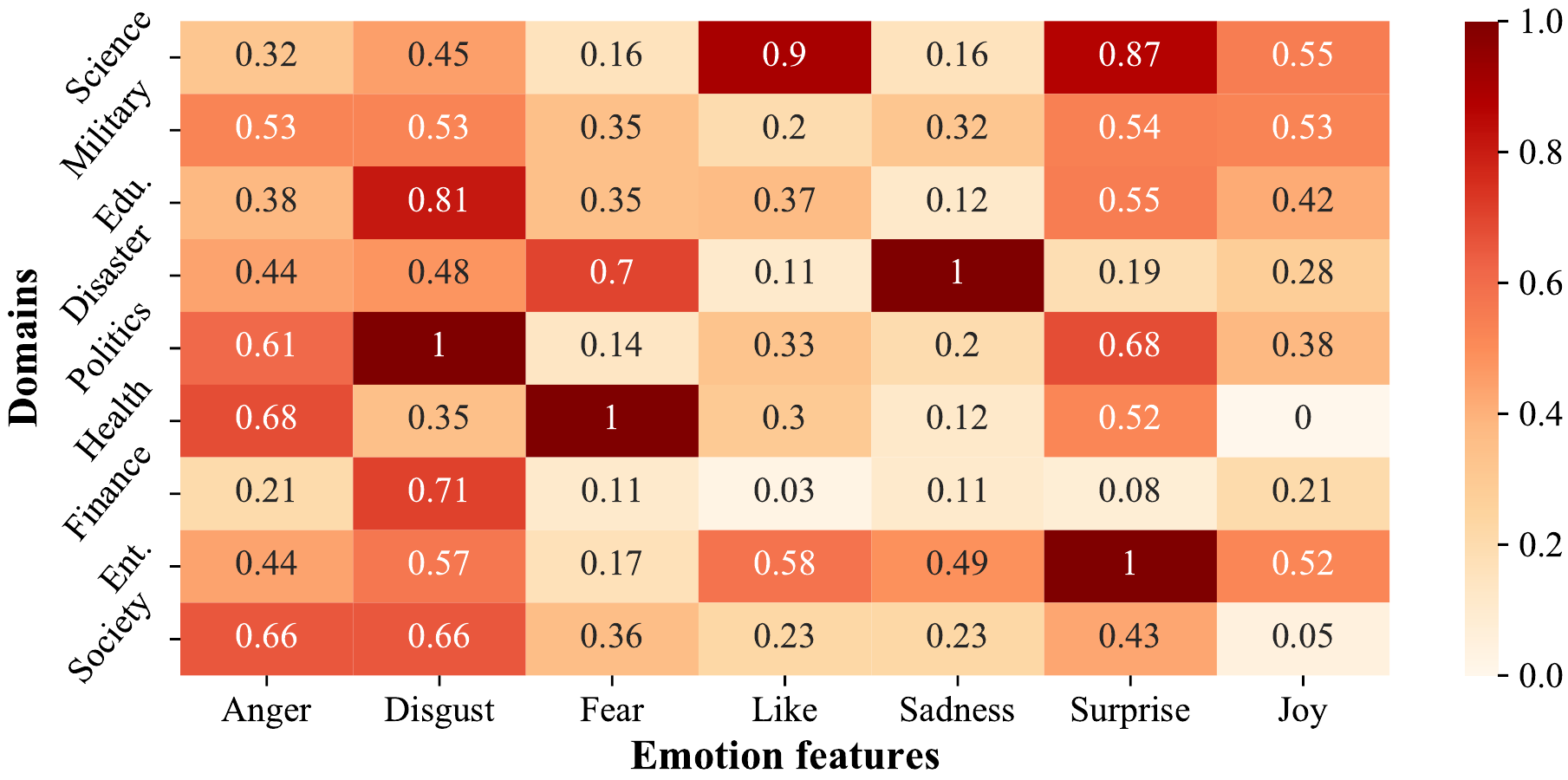}
\label{fig:analysis_b}
}
\subfigure[Style features of real news]{
\centering
\includegraphics[width=0.5\columnwidth]{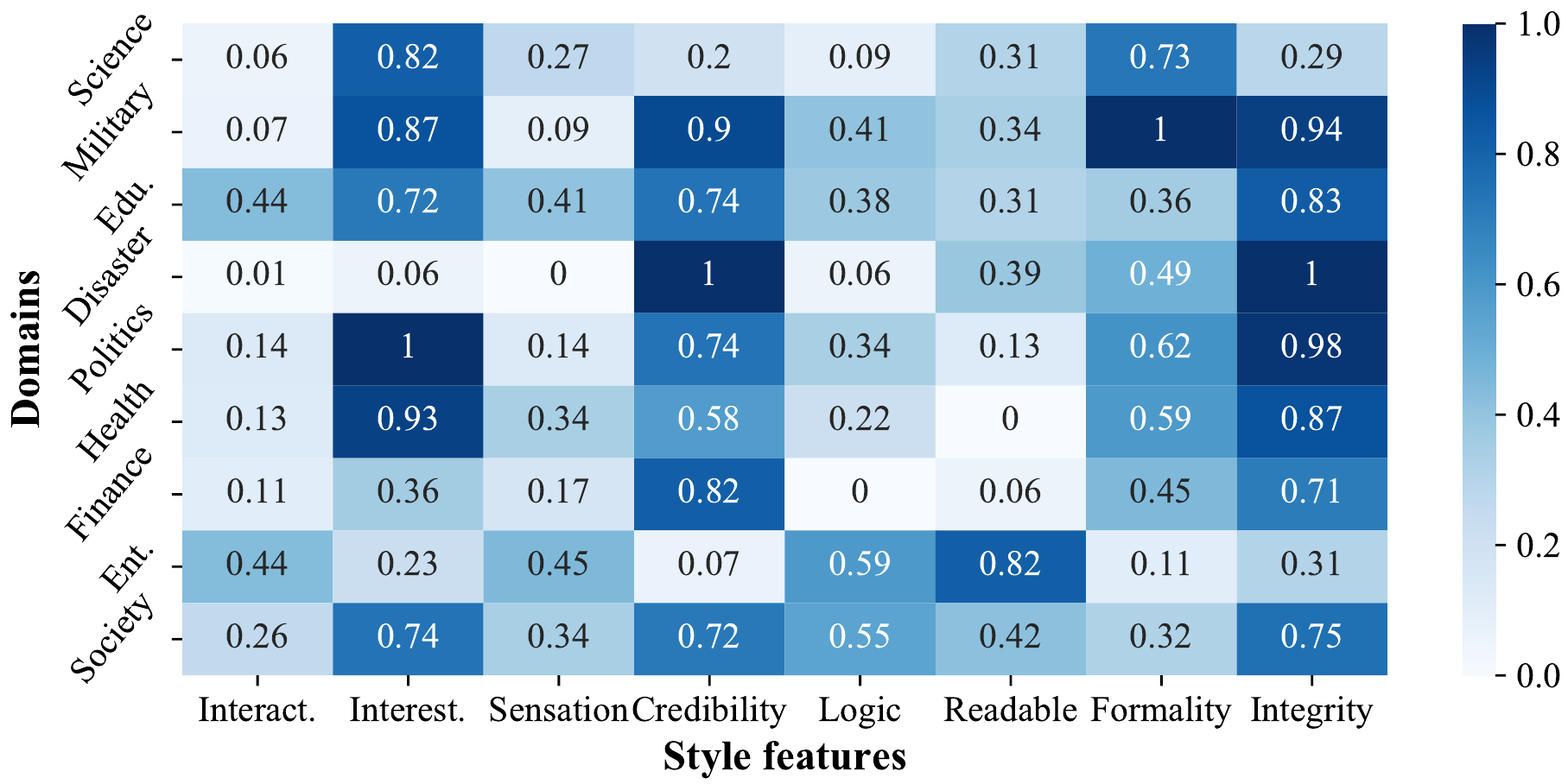}
\label{fig:analysis_d}
}
\subfigure[Emotion features of real news]{
\centering
\includegraphics[width=0.5\columnwidth]{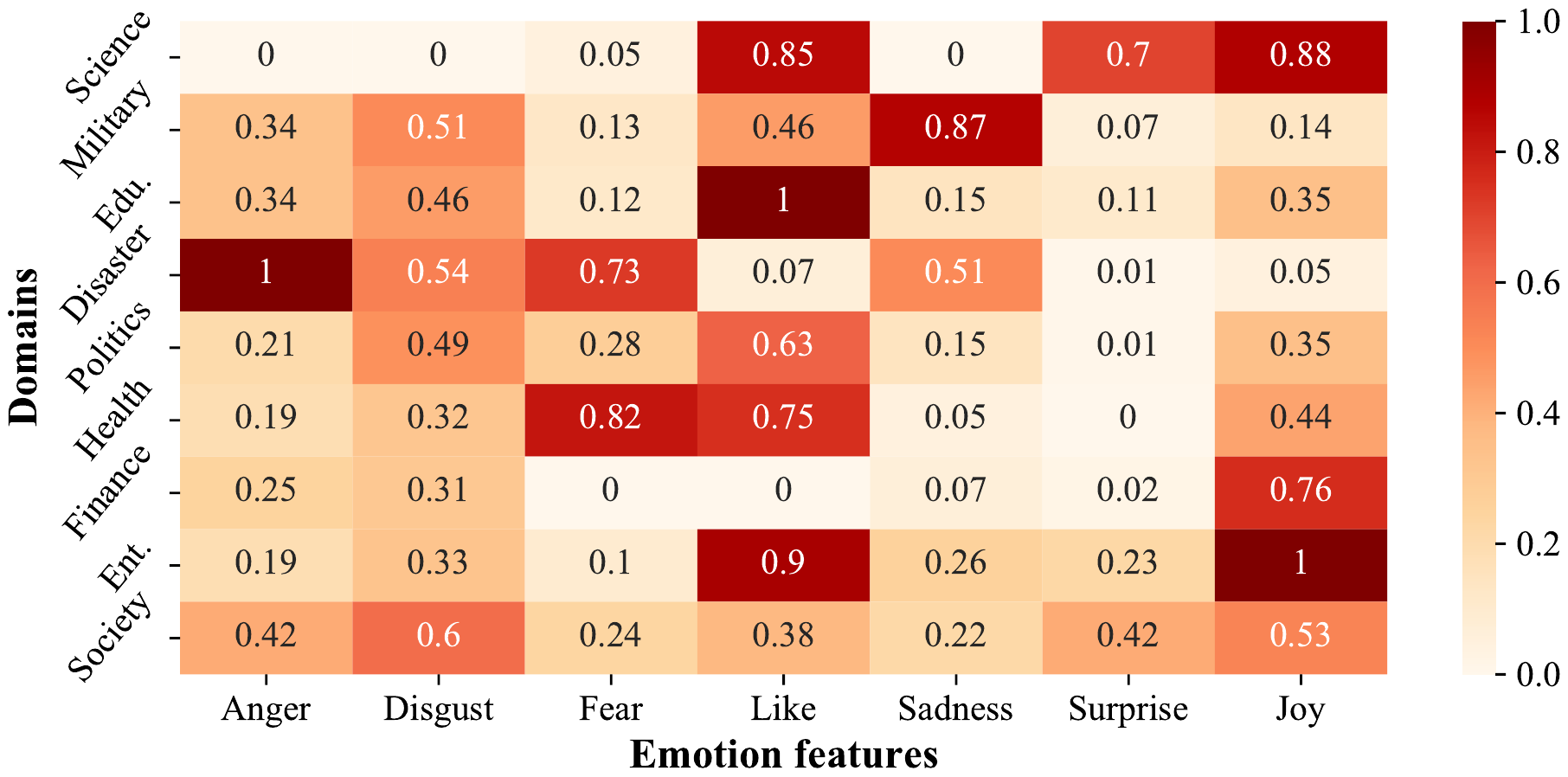}
\label{fig:analysis_e}
}
\end{minipage}
\caption{Figure (a) denotes the Top 20 words in the nine domains. Figures (b)-(e) are heatmaps with the average value of different features in the nine domains. Figures (b) and (d) denote style features. Figures (c) and (e) denote emotion features. In fake news (the first row), there are distinct style and emotion features from real news (the second row).}
\label{fig:analysis_domainshift}
\vspace{-0.2cm}
\end{figure*}

In this paper, we first investigate multi-domain news data and find that multi-domain fake news detection empirically faces two challenges (see detailed analysis in Section~\ref{sec:analysis}):

\textbf{\textit{(1) Domain shift among multiple news domains.}} 
Various news domains have significant domain discrepancies, e.g., words, emotions, styles. Figure~\ref{fig:intro} shows that the topics of Politics and Health Domains are different.
In addition, as shown in Figure~\ref{fig:analysis_domainshift}, we find distributions of the writing styles, word usages, publisher emotions of various domains would be largely different, and the differences among domains are called domain shift~\cite{pan2009survey}. Generally, domain shift could seriously influence the effectiveness of joint training multi-domain data~\cite{ganin2016domain,peng2019moment,zhu2019aligning}. Thus, it is essential to propose well-designed multi-domain models to alleviate the influence of domain shift.

\textbf{\textit{(2) Domain labeling incompleteness for news pieces.}} In a real-world news platform, a news piece is released in a single domain (channel). However, a news piece is a mixture of diverse elements which makes it relate to multiple news domains. As shown in Figure~\ref{fig:intro}, news (c) is categorized into the Politics Domain, while it also involves a health topic on COVID-19 as shown in news (b). In addition, in Section~\ref{sec:analysis}, we find the domain boundary of news domains is not clear, which also indicates a news piece could have multiple domain labels. Since domain labels are useful for multi-domain learning~\cite{pan2009survey,zhuang2020comprehensive}, completing the domain labels is important for building an accurate multi-domain fake news detection system.

For multi-domain fake news detection, \cite{silva2021embracing} combined two existing datasets FakeNewsNet~\cite{shu2020fakenewsnet} and CoAID~\cite{cui2020coaid} into a three-domain dataset and proposed a method to preserve domain-specific and domain-shared knowledge. In practice, a news platform consists of far more than three domains, which indicates more challenging domain shift problem, and learning domain-shared knowledge is proved difficult under the challenging scenario~\cite{ma2018modeling,zhu2019aligning}. Nan et al.~\cite{nan2021mdfend} proposed a dataset named Weibo21 with nine domains for multi-domain fake news detection and a simple baseline based on Mixture-of-Experts~\cite{jacobs1991adaptive,ma2018modeling}. However, Nan et al.~\cite{nan2021mdfend} ignored the problem of domain labeling incompleteness.



Along this line, we propose a novel Memory-guided Multi-view Multi-domain Fake News Detection Framework (M$^3$FEND). Since the distributions of word usage, style, and emotion vary from domain to domain, we firstly propose three multi-channel networks to model news pieces from semantic view, emotional view, and stylistic view, named SemNet, EmoNet, and StyNet, where each channel can focus on different patterns. Cross-view interactions could capture associations among different views and produce more diverse combinations of views which benefits modeling the domain discrepancy~\cite{li2018survey,zadeh2018memory,hassani2020contrastive,wu2022multi,li2022user}. With the advantage of cross-view interactions, we propose a Multi-head Adaptive Cross-view Interactor to adaptively learn various cross-view interactions. Note that the discriminability of views varies from domain to domain, e.g., the style view is discriminative for Science Domain while not for Entertainment Domain as shown in Section~\ref{sec:analysis}. Therefore, we propose a Domain Adapter with domain information as input to adaptively aggregate discriminative cross-view representations for news in different domains.

To complete domain labels and enrich domain information, we propose a Domain Memory Bank which consists of a Domain Characteristics Memory and multiple Domain Event Memories. Domain Characteristics Memory aims to automatically capture and store information of domain characteristics. In addition, each domain has a Domain Event Memory matrix which records all news released in this domain. Each Domain Event Memory matrix consists of several memory units, and each unit represents a cluster of similar news. Then, we compute the similarity between each Domain Event Memory matrix and a certain news piece. The similarity can denote the distribution of domain labels, and we utilize the similarity distribution to enrich domain information for the news piece. The enriched domain information is utilized to guide the Domain Adapter to aggregate cross-view representations.



M$^3$FEND has been successfully deployed in an online fake news detection system that handles millions of news pieces every day, leading to a more trustful online news ecosystem. Our contributions are summarized as follows:
\begin{enumerate}
    \item We investigate the problem of multi-domain fake news detection and point out two challenges of domain shift and domain labeling incompleteness.
    \item To solve the challenges, we propose a novel Memory-guided Multi-view Multi-domain Fake News Detection Framework (M$^3$FEND), which can improve the detection performance of most domains.
    \item We conduct both offline and online experiments to demonstrate the effectiveness of M$^3$FEND. Our code is available at \url{https://github.com/ICTMCG/M3FEND}.
\end{enumerate}

\vspace{-0.3cm}\section{Analysis}\label{sec:analysis}
In this section, we first investigate domain shift in multi-domain fake news detection. We analyze domain shift from three views, including word usage, writing style, publisher emotion. In addition, we testify to the problem of domain labeling incompleteness. All analysis is performed based on a Chinese multi-domain fake news detection dataset~\cite{nan2021mdfend} with nine domains, denoted as Ch-9 in this paper. The statistics of the Ch-9 dataset
are shown in Table~\ref{tab:data_statistic_ch}.

\vspace{-0.3cm}
\subsection{Domain Shift}\label{sec:domain_shift}
\subsubsection{Word Usage.} Intuitively, news pieces published in different domains have various topics and domain-specific word usage~\cite{silva2021embracing,nan2021mdfend}. To testify the domain-specific word usage, we plot word clouds of different domains based on word frequency in Figure~\ref{fig:word_usage}. From Figure~\ref{fig:word_usage}, we observe that different news domains have significant differences in the frequently used words. 

\subsubsection{Writing Style.} Fake news publishers utilize particular writing styles to appeal to and persuade a wide scope of consumers. The writing styles of various domains could be largely different. Based on~\cite{yang2019write} that mines writing style for news on the social network, we extract eight writing styles including Readability, Logic, Credibility, Formality, Interactivity, Interestingness, Sensation, and Integrity. The average normalized writing style features of fake and real news are shown in Figure~\ref{fig:analysis_a} and~\ref{fig:analysis_d}, respectively. We can find that style features of fake and real news pieces are different, and the discriminative style features of various domains are also different, e.g., the Logic feature is beneficial to detect fake news for Science Domain while not for Finance Domain.

\subsubsection{Publisher Emotion.} Fake news is more likely to be emotional or inflammatory than real news~\cite{ajao2019sentiment,shu2017fake}, and emotional signals are useful patterns to detect fake news~\cite{giachanou2019leveraging,zhang2021mining}. According to~\cite{zhang2021mining}, we evaluate seven kinds of emotions in publisher contents with an affective lexicon ontology database~\cite{xu2008constructing}, including Anger, Disgust, Fear, Sadness, Surprise, Like, Joy. The emotional features of fake and real news are shown in Figure~\ref{fig:analysis_b} and~\ref{fig:analysis_e}. We can observe that the discriminative emotion features of various domains are different, e.g., fake news publishers of Politics Domain show more disgust, while the publishers of Health Domain show more anger.


\begin{figure}[t!]
\centering
\includegraphics[width=1\columnwidth]{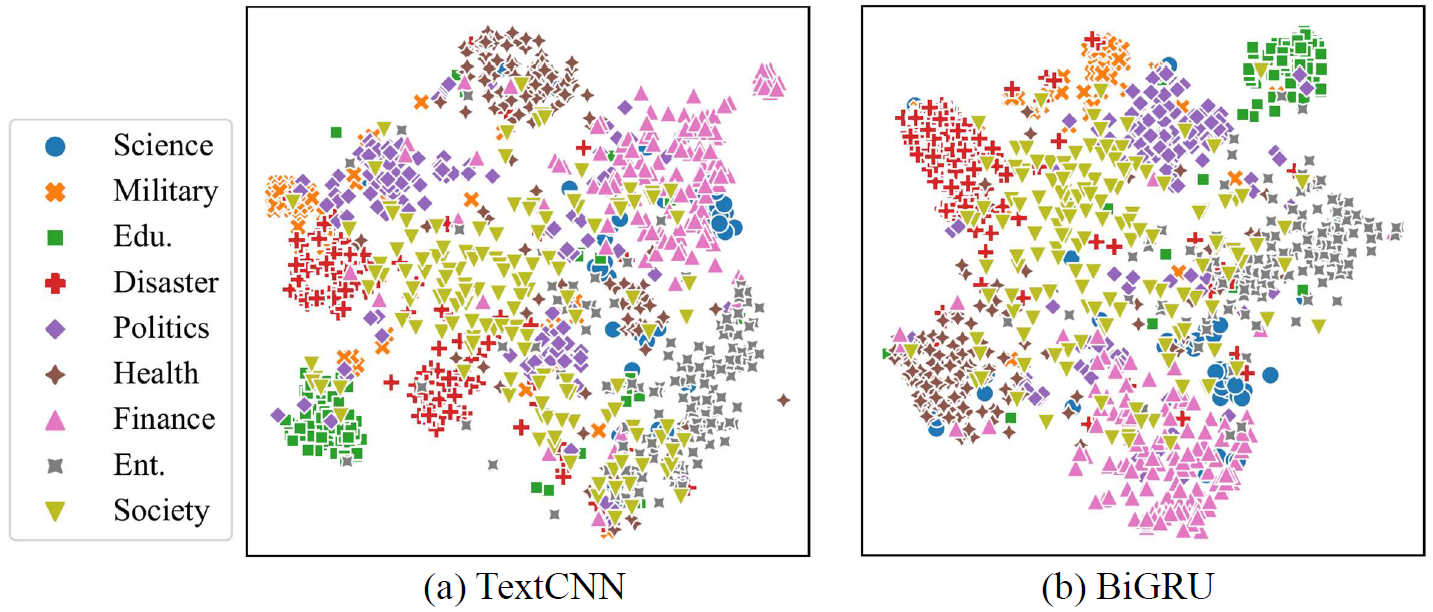}
\vspace{-0.5cm}
\caption{Visualizations of unclear domain boundaries with the domain classification task using t-SNE on the Ch-9.}
\label{fig:analysis_tsne}
\end{figure}



\begin{table}[t]
\setlength{\abovecaptionskip}{3pt}
\setlength{\belowcaptionskip}{5pt}
  \centering
  \caption{Accuracy on Ch-9 for domain classification.}
  \setlength{\tabcolsep}{0.5pt}
    \begin{tabular}{cccccccccc}
    \toprule
          & \rotatebox{40}{Science} & \rotatebox{40}{Military} & \rotatebox{40}{Edu.} & \rotatebox{40}{Disaster} & \rotatebox{40}{Politics} & \rotatebox{40}{Health} & \rotatebox{40}{Finance} & \rotatebox{40}{Ent.} & \rotatebox{40}{Society} \\
    \midrule
    TextCNN & 0.42  & 0.57  & 0.75  & 0.74  & 0.77  & 0.82  & 0.83  & 0.84  & 0.78  \\
    BiGRU & 0.56  & 0.62  & 0.87  & 0.69  & 0.77  & 0.78  & 0.84  & 0.77  & 0.79  \\
    \bottomrule
    \end{tabular}\label{tab:error}
\end{table}%

\vspace{-0.3cm}
\subsection{Domain Labeling Incompleteness}
Generally, a news piece is manually categorized into a single domain, but the content of a news piece could be related to several domains at the same time. In other words, it is difficult to discriminate domains by news content. To testify this problem, we conduct a supervised domain classification task, which exploits deep models (TextCNN~\cite{kim-2014-convolutional}, BiGRU~\cite{ma2016detecting}) to predict domain labels by news content. Firstly, We visualize in Figure~\ref{fig:analysis_tsne} the network activations of the test set (before classifier) learned by TextCNN and BiGRU learned using t-SNE embeddings~\cite{donahue2014decaf}. From Figure~\ref{fig:analysis_tsne}, we can find that the domain decision boundary is not clear, and many samples of different domains are overlapping, especially Society Domain. For fake news detection without the object of domain classification, it is more difficult to discriminate domain labels. In addition, we evaluate and list the accuracy of domain classification tasks in Table~\ref{tab:error}, and the unsatisfying accuracy also demonstrates the content of the news piece could be related to several domains. Thus, domain labeling incompleteness is an important issue for multi-domain fake news detection.

\section{Model}
\begin{figure*}[ht]
\setlength{\abovecaptionskip}{3pt}
\setlength{\belowcaptionskip}{-11pt}
\centering
\begin{minipage}[b]{1\linewidth}
\centering
\includegraphics[width=0.88\linewidth]{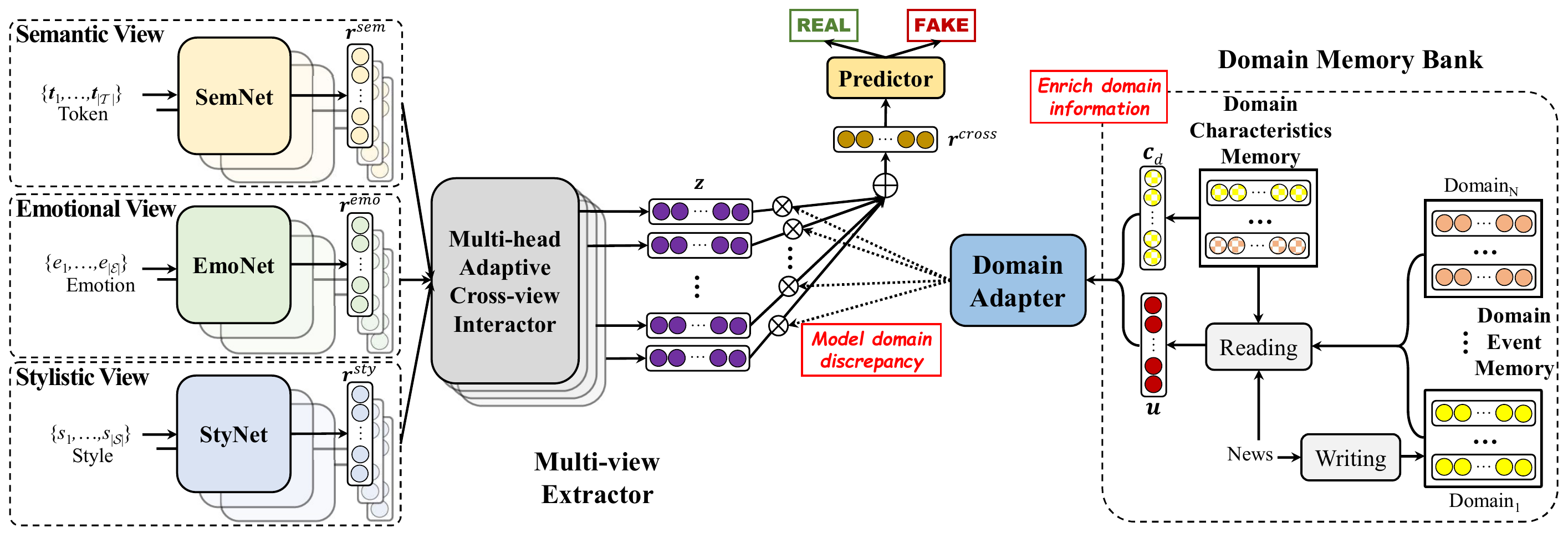}
\end{minipage}
\caption{Overall architecture of the Memory-guided Multi-view Multi-domain Fake News Detection Framework (M$^3$FEND). The model consists of a Multi-view Extractor, a Domain Memory Bank, a Domain Adapter, and a Predictor. The Multi-view Extractor aims to extract multi-view representations and model cross-view interactions. The Domain Memory Bank stores and provides enriched domain information. The Domain Adapter aggregates discriminative cross-view representations for news in different domains. The Predictor uses the aggregated representations for the final prediction.}\label{fig:network}
\end{figure*}

In this section, we introduce the proposed Memory-guided Multi-view Multi-domain Fake News Detection Framework (M$^3$FEND) in detail, and the overall structure is shown in Figure~\ref{fig:network}. We first give the definitions of multi-domain fake news detection. Then, we detail the components of M$^3$FEND: The Multi-view Extractor extracts multi-view representations and adaptively models cross-view interactions. The Domain Memory Bank tackles the problem of domain label incompleteness and enriches domain information. With enriched domain information as input, the Domain Adapter models the domain discrepancy and feeds useful cross-view representations into the Predictor.

\subsection{Problem Definition}
Let $P$ be a news piece on social media, and the text of the news piece consists of $T$ tokens (words). We adopt RoBERTa~\cite{liu2019roberta,cui2019pre}, a robustly optimized BERT~\cite{devlin2019bert} pre-training model to encode tokens of the news content as $\mathcal{T} = \{\bm{t}_1, ..., \bm{t}_{|\mathcal{T}|}\}$, where $\bm{t} \in \mathbb{R}^O$ denotes an embedding and $O$ indicates the dimension of embeddings. Inspired by~\cite{zhang2021mining} and the above analysis, we extract emotion features from the news piece, including emotion category, emotional intensity, sentiment score, and so on. The emotion features are denoted as $\mathcal{E}=\{e_1, ..., e_{|\mathcal{E}|}\}$. Similarly, we extract style features based on~\cite{yang2019write}, and the writing style features of the news piece are denoted as $\mathcal{S}=\{s_1, ..., s_{|\mathcal{S}|}\}$. Both emotion and style features are numerical features~\cite{castillo2011information}. Each news piece has a ground-truth label $y \in \{0, 1\}$, where $1$ and $0$ denote the new piece is fake and real, respectively. In addition, each news piece is manually categorized into a single domain with a domain label $d \in \{Domain_1, ..., Domain_N\}$, where $N$ indicates the number of domains. Given a news piece $P$ and a domain label $d$, multi-domain fake news detection aims to detect whether the news is fake or real.

\subsection{Multi-view Extractor}

According to the above analysis in Section~\ref{sec:domain_shift}, various views can be useful to detect fake news. Thus, firstly, it is necessary to extract news representations from multiple views. Specifically, we exploit three deep extractors to extract news representations from the semantic view, emotional view, and stylistic view, respectively.


\textit{Semantic Network (SemNet):} It aims to extract representations from the semantic view with the text content of news pieces. In this paper, TextCNN~\cite{kim-2014-convolutional} is employed as SemNet with encoded embeddings $\{\bm{t}_1, ..., \bm{t}_{|\mathcal{T}|}\}$ as input, which can be formulated as:
\begin{equation}
    \bm{r}^{sem} = \text{SemNet}(\{\bm{t}_1, ..., \bm{t}_{|\mathcal{T}|}\}).
\end{equation}

\textit{Emotion Network (EmoNet):} This part focuses on modeling the emotional view with emotional signals $\{e_1, ..., e_{|\mathcal{E}|}\}$. In this paper, Multilayer Perceptron (MLP) is adopted as EmoNet to extract emotional representations of news pieces $\bm{r}^{emo}$. Specifically, an EmoNet denoted as:
\begin{equation}
    \bm{r}^{emo} = \text{EmoNet}(\{e_1, ..., e_{|\mathcal{E}|}\}).
\end{equation}

\textit{Style Network (StyNet):} It pays attention to the stylistic view with writing style features $\{s_1, ..., s_{|\mathcal{S}|}\}$. MLP is employed as StyNet to extract style representations $\bm{r}^{sty}$, which can be formulated as:
\begin{equation}
    \bm{r}^{sty} = \text{StyNet}(\{s_1, ..., s_{|\mathcal{S}|}\}).
\end{equation}

Inspired by multi-channel CNN~\cite{he2016deep,huang2017densely}, it is beneficial to extract multiple representations for each view with different, learned multi-channel SemNet, EmoNet, and StyNet. Multi-channel extractors allow the model to jointly attend to information from different representation subspaces~\cite{he2016deep,vaswani2017attention}, and different representations could focus on various patterns~\cite{tao2019multi,zhu2019multi}.  Then, we could obtain three groups of representations, each of which corresponds to semantics, emotion, and style views, respectively, denoted as $\{\bm{r}^{sem}_i\}_{i=1}^{k_{sem}}$, $\{\bm{r}^{emo}_i\}_{i=1}^{k_{emo}}$, and $\{\bm{r}^{sty}_i\}_{i=1}^{k_{sty}}$, where $k_{sem}$, $k_{emo}$, and $k_{sty}$ indicate the channel number of SemNet, EmoNet, and StyNet, respectively.


\textit{Multi-head Adaptive Cross-view Interactor:} Cross-view interactions could capture associations among different views and produce more diverse combinations of views. A direct way is to enumerate the view combinations with all views, which will result in high computational complexity. In addition, such the method cannot model the importance of views in a cross-view interaction, and indiscriminative views may lead to noisy view combinations that degrade model performance. To tackle the problems, we propose an Adaptive Cross-view Interactor to automatically learn cross-view representations, which is formulated as:
\begin{equation}
\begin{split}
    \bm{z} = \exp & \left [  \sum_{i=1}^{k_{sem}} a_i^{sem} \ln \bm{r}^{sem}_i + \sum_{j=1}^{k_{emo}} a_j^{emo} \ln \bm{r}^{emo}_j \right . \\
    & \left . + \sum_{q=1}^{k_{sty}} a_q^{sty}\ln{\bm{r}^{sty}_q} \right ] ,
\label{eq:cross_feature}
\end{split}
\end{equation}
where $a^{sem}$, $a^{emo}$, and $a^{sty}$ indicate the importance of different representations from semantic, emotional, and stylistic views, respectively, which are learnable parameters. $\bm{z}$ indicates a representation of a cross-view interaction. In addition, the cross-view interaction can be denoted as product operation of various views, and the Equation~\ref{eq:cross_feature} can be reformulated as:
\begin{equation}
    \bm{z} = \prod_{i=1}^{k_{sem}} (\bm{r}_i^{sem})^{a_i^{sem}} \odot \prod_{j=1}^{k_{emo}} (\bm{r}_j^{emo})^{a_j^{emo}} \odot \prod_{q=1}^{k_{sty}} (\bm{r}_q^{sty})^{a_q^{sty}},
\end{equation}
where $\odot$ denotes element-wise product operation and $\prod$ indicates continuous element-wise product. An Adaptive Cross-view Interactor can extract representations of a cross-view interaction. However, a single cross-view representation is not discriminative for all domains.  For example, the Health Domain may largely rely on the semantic and emotional views, while the Science Domain depends on the semantic and stylistic views. Thus, it is necessary to extract various cross-view representations. Along this line, to model different cross-view interactions, we propose a Multi-head Adaptive Cross-view Interactor with $H$ heads, and each head adaptively learns a kind of cross-view representation. $H$ cross-view representations are denoted as $\{\bm{z}_i\}_{i=1}^H$.


\subsection{Domain Memory Bank}
Domain Memory Bank aims to complete domain labels and enrich domain information in news pieces, which consists of a Domain Characteristics Memory and a set of Domain Event Memories. The enriched domain information is utilized as the input of the Domain Adapter.

\subsubsection{Domain Characteristics Memory.} It aims to automatically capture and store domain characteristics. Domain Characteristics Memory can be denoted as $\mathcal{C} = \{\bm{c}_i\}_{i=1}^N$, where $\bm{c}_i$ represents a memory unit of the $i$-th domain and $N$ denotes the number of domains. All parameters of the Domain Memory $\mathcal{C}$ are randomly initialized. The memory unit $\bm{c}_i$ is only learned from training samples of the $i$-th domain, so it could be seen as the characteristics representation of the $i$-th domain. 

\subsubsection{Domain Event Memory.} A certain news piece is given a specific domain label $d$, but the news piece may simultaneously contain information of other domains. To tackle the problem of domain labeling incompleteness, we propose a Domain Event Memory mechanism that aims to discover potential domain labels of news and enrich domain information.
The key idea is that a Domain Event Memory matrix records all news released in this domain, and for a certain news piece, we evaluate the similarity between the news and all Domain Event Memory matrices. The similarity can represent the distributions of potential domain labels.

Domain Event Memory of the $j$-th domain is denoted as $\mathcal{M}_j = \{\bm{m}_i\}_{i=1}^Q$, where $\bm{m}$ represents a memory unit and $Q$ denotes the number of memory units. A memory unit $\bm{m}$ represents a set of similar news pieces, and all news pieces in a specific news domain can be divided into $Q$ clusters. Each domain has a Domain Event Memory matrix, so there are $N$ Domain Event Memory matrices.

\textit{Initialization.} We initialize $\bm{m}$ by clustering similar news pieces with K-means algorithm. A news piece is represented as $\bm{n} = [\mathcal{G}(\{\bm{t}_1, ..., \bm{t}_{|\mathcal{T}|}\});\{e_1, ..., e_{|\mathcal{E}|}\};\{s_1, ..., s_{|\mathcal{S}|}\}] \in \mathbb{R}^{I}$, 
where $\mathcal{G}(\cdot)$ denotes a learnable attention layer with a mask operation (replacing the padding position as $-\inf$) following~\cite{nan2021mdfend}.
Before training, we obtain representations of all news pieces, and we aggregate news representations into $Q$ clusters using K-means for each domain, respectively. For a specific domain, all centers of clusters are utilized to initialize its memory units.

\textit{Reading operation.} This operation aims to evaluate the similarity between a given news piece and all Domain Event Memory matrices. Specifically, for a given news piece, we first find all similar memory units $\bm{m}$ in a certain Domain Event Memory matrix $\mathcal{M}_j$ and aggregate them into a domain representation:
\begin{equation}
    \bm{o}_j = \text{softmax}(\bm{n}Wg(\mathcal{M}_j)/\tau )\mathcal{M}_j,
\end{equation}
where $g(\cdot)$ denotes transpose function, and $W \in \mathbb{R}^{I\times I}$ is a learnable parameter matrix. We set $\tau = 0.01$ to only find the most similar event cluster. All domain representations are concentrated into a matrix as $\mathcal{D} = [\bm{o}_1, ..., \bm{o}_N] \in \mathbb{R}^{N \times I}$, where $N$ indicates the number of domains. Then, the similarity distribution can be denoted as:
\begin{equation}
    \bm{v} = \text{softmax}(\bm{n}Vg(\mathcal{D})),\label{eq:event_domain_memory}
\end{equation}
where $V \in \mathbb{R}^{I\times I}$ is a learnable parameter matrix, and $\bm{v} \in \mathbb{R}^N$ indicates the similarity distribution.

For a certain news piece, according to the domain label $d$, we look up the Domain Characteristics Memory $\mathcal{C}$ to obtain an explicit domain representation $c_d$. Then, with the similarity distribution $\bm{v}$, we evaluate an implicit domain representation as: $\bm{u} = \sum_{i=1}^o v_i \bm{c}_i$,
where $\bm{c}_i$ is the representation of $i$-th domain and $\bm{v}_i$ denotes the $i$-th element of the similarity distribution vector. The implicit domain representation contain potential domain information. Finally, the implicit representation $\bm{u}$ and the explicit representation $\bm{c}_d$ are concentrated into $[\bm{c}_d, \bm{u}]$ to represent enriched domain information in the news piece.

\textit{Writing operation.} A given domain label $d$ indicates that the news piece contains topics of a certain domain. Thus, we store the news piece in the specific Domain Event Memory $\mathcal{M}_d$. Inspired by neural turning machine (NTM)~\cite{graves2014neural,sheng2021article}, when writing the Domain Event Memory matrix $\mathcal{M}_d$, it will be erased first before new information is added. We compute the similarity between the news piece and each memory unit as $\bm{sim} = \text{softmax}(\bm{n}Wg(\mathcal{M}_d) / \tau)$. The erasing vector from the memory unit $\bm{m}_i$ is denoted as: $\bm{erase}_i = \bm{sim}_i \cdot \bm{m}_i$. Then, an adding vector for the memory unit $\bm{m}_i$ is denoted as $\bm{add}_i = \bm{sim}_i \cdot \bm{n}$. The overall writing operation can be formulated as:
\begin{equation}
    \bm{m}_i = \bm{m}_i - \beta \bm{erase}_i + \beta \bm{add}_i,
\end{equation}
where $\beta$ is a parameter to control that the proportions of memory erasing and adding are the same (here, $0.05$). 

\subsubsection{Discussion}
In this work, we utilize the Domain Memory Bank to complete the domain labels. Note that domain labeling incompleteness is different from the missing label issue. Domain labeling incompleteness indicates a news piece is given a single domain label, but it could be related to multiple news domains, while the missing label issue denotes some samples have no label. Pseudo label generation has been widely used in semi-supervised scenarios, which trains a classifier to infer the missing labels~\cite{berthelot2019mixmatch,zhang2021flexmatch}. Intuitively, such pseudo label generation methods can be exploited to complete domain labels. However, in multi-domain fake news detection, it is difficult to train a satisfying domain classifier as shown in Section~\ref{sec:analysis}. Thus, the pseudo labels generated by the domain classifier are unreliable, and the model might be misled by the wrong domain labels. To achieve a similar function and avoid the above issues, we propose the Domain Memory Bank. Different from pseudo label generation methods, our design is not determined by the accuracy of the Domain Memory Bank as it is trained end-to-end with the fake news detection as the final objective. In addition, our method has better transparency because it preserves more event information to help us know how it finds potential domain labels as shown in Table~\ref{tab:case}.

\subsection{Domain Adapter}
Due to the existing domain discrepancy, the discriminative cross-view representations of various domains could be different. Thus, we propose a Domain Adapter to model the domain discrepancy. The Domain Adapter takes the enriched domain representation $[\bm{c}_d, \bm{u}]$ from the Domain Memory Bank as input to aggregate useful cross-view representations for final prediction. Specifically, the aggregated cross-view representation is formulated as:
\begin{equation}
    \bm{r} = \sum_{i = 1}^{H} w_i \bm{z}_i, \quad \bm{w} = \text{softmax} (f([\bm{c}_d, \bm{u}])),\label{eq:cross_gate}
\end{equation}
where $f(\cdot)$ is a feed-forward network and $\bm{w} \in \mathbb{R}^{H}$ is the weight vector representing the importance of different cross-view representations. $H$ is the number of cross-view representations. Note that the weight vector $\bm{w}$ could help us understand the decision process of a certain news piece depends on which cross-view interactions.

\begin{table}[t]
  \centering
  \setlength{\abovecaptionskip}{-0.010cm}
  \caption{Data Statistics of Ch-9}
    \begin{tabular}{lrrrrr}
    \toprule
    \textbf{Domain} & \textbf{Science} & \textbf{Military} & \textbf{Edu.} & \textbf{Disasters} & \textbf{Politics} \\
    \midrule
    \textbf{\#Real} & 143   & 121   & 243   & 185   & 306 \\
    \textbf{\#Fake} & 93    & 222   & 248   & 591   & 546 \\
    \midrule
    \textbf{Total} & 236   & 343   & 491   & 776   & 852 \\
    \midrule
    \midrule
    \textbf{Domain} & \textbf{Health} & \textbf{Finance} & \textbf{Ent.} & \textbf{Society} & \textbf{All} \\
    \midrule
    \textbf{\#Real} & 485   & 959   & 1,000  & 1,198  & 4,640 \\
    \textbf{\#Fake} & 515   & 362   & 440   & 1,471  & 4,488 \\
    \midrule
    \textbf{Total} & 1,000  & 1,321  & 1,440  & 2,669  & 9,128 \\
    \bottomrule
    \end{tabular}%
  \label{tab:data_statistic_ch}%
\end{table}%

\begin{table}[t]
  \centering
  \setlength{\abovecaptionskip}{-0.010cm}
  \caption{Data Statistics of En-3}
    \begin{tabular}{lrrrr}
    \toprule
    \textbf{Domain} & \textbf{GossipCop} & \textbf{PolitiFact} & \textbf{COVID}  & \textbf{All}  \\
    \midrule
    \textbf{\#Real} & 16,804   & 447   & 4,750 & 22,001  \\
    \textbf{\#Fake} & 5,067    & 379   & 1,317 & 6,763  \\
    \midrule
    \textbf{Total} & 21,871   & 826   & 6,067 & 28,764 \\
    \bottomrule
    \end{tabular}%
  \label{tab:data_statistic_en}%
\end{table}%

\subsection{Predictor}
With the aggregated cross-view representation $\bm{r}$, we predict the probability of a news piece $P$ being fake with:
\begin{equation}
    \hat{p} = \text{Sigmoid}(\text{MLP}(\bm{r})).
\end{equation}

All parameters are learnable and can be optimized by minimizing the cross entropy loss with back-propagation:
\begin{equation}
    \mathcal{L} = -y \log \hat{p} - (1-y) \log (1 - \hat{p}).
\end{equation}

\section{Experiments}\label{sec:experiment}

\begin{table}[t]
\setlength{\abovecaptionskip}{3pt}
\setlength{\belowcaptionskip}{5pt}
  \centering
  \caption{Emotion Features.}
    \begin{tabular}{cp{17em}}
    \toprule
    \textbf{Feature} & \multicolumn{1}{c}{\textbf{Description}} \\
    \midrule
    \multirow{3}[1]*{\shortstack{Emotional \\Category} } & The probabilities that the given text contains certain emotions obtained from publicly available emotion classifiers. \\
    \midrule
    \multirow{3}[1]*{\shortstack{Emotional \\Lexicon} } & The overall emotion score that is aggregated from scores of each word and the whole text across all the emotions. \\
    \midrule
    \multirow{4}[2]*{\shortstack{Emotional \\Intensity} } & The overall intensity scores which is extracted from the existing emotion dictionaries annotated with similar process as Emotional Lexicon. \\
    \midrule
    \multirow{3}[4]*{\shortstack{Sentiment \\Score} } & The degree of the positive or negative polarity of the whole text calculated by using sentiment dictionaries or public toolkits. \\
    \midrule
    \multirow{3}[1]*{\shortstack{Auxiliary \\Features} } & The frequency of emoticons, punctuations, sentimental words, personal pronoun, and uppercase letters. \\
    \bottomrule
    \end{tabular}%
  \label{tab:emofea}%
\end{table}%

In this section, we conduct experiments with the aim of answering the following research questions:
\begin{itemize}
\setlength{\itemindent}{6pt}
\setlength{\parsep}{0pt}
\setlength{\parskip}{0pt}
    \item[\textbf{RQ1}] Does our proposed M$^3$FEND outperform other approaches in different datasets?
    \item[\textbf{RQ2}] Can M$^3$FEND framework improve the performance of a real-world fake news detection system?
    \item[\textbf{RQ3}] What are the effects of different views and components in our proposed M$^3$FEND?
    \item[\textbf{RQ4}] How does M$^3$FEND model the domain discrepancy and find potential domain labels? How sensitive are the hyperparameters?
\end{itemize}

\subsection{Experimental Settings}
\subsubsection{Datasets.} We evaluate M$^3$FEND with baselines on both English and Chinese datasets of multi-domain fake news detection tasks. The statistics of the Ch-9 and En-3 datasets are shown in Table~\ref{tab:data_statistic_ch} and \ref{tab:data_statistic_en}, respectively.

\textit{English Dataset}. Following~\cite{silva2021embracing}, we combine FakeNewsNet~\cite{shu2020fakenewsnet} and COVID~\cite{li2020mm} into an English multi-domain fake news detection dataset with three domains, including GossipCop, PolitiFact, and COVID, which we named as En-3.

\textit{Chinese Dataset}~\cite{nan2021mdfend}. It is a Chinese multi-domain fake news detection dataset collected from Sina Weibo with nine domains, including Science, Military, Education, Disaster, Politics, Health, Finance, Entertainment, and Society. We denote the full dataset as Ch-9. In addition, to testify the effectiveness of M$^3$FEND under various scenarios, we sample two datasets as Ch-3 and Ch-6. Ch-3 contains the same three domains as the En-3 dataset. Ch-6 contains 6 domains that are related to daily life, including Education, Disaster, Health, Finance, Entertainment, and Society.

\begin{table*}[t]
\setlength{\abovecaptionskip}{3pt}
\setlength{\belowcaptionskip}{5pt}
\begin{minipage}{.5\linewidth}
\setlength{\tabcolsep}{1.5pt}
  \centering
  \caption{Results on the En-3 dataset.  * ($p \le 0.05$) and ** ($p \le 0.005$) indicate paired t-test of M$^3$FEND vs. the best baseline.}
    \begin{tabular}{cc||ccc||ccc}
    \toprule
    \multirow{2}[4]{*}{} & \multirow{2}[1]{*}{Method} & \multirow{2}[1]{*}{Gossip.} & \multirow{2}[1]{*}{Polit.} & \multirow{2}[1]{*}{COVID} & \multicolumn{3}{c}{overall} \\
\cline{6-8}          &       &       &       &       & F1    & Acc   & AUC \\
    \midrule
    \multicolumn{1}{c}{\multirow{3}[2]{*}{\begin{sideways}single\end{sideways}}} & BiGRU & 0.7666  & 0.7722  & 0.8885  & 0.7958  & 0.8668  & 0.8840  \\
          & TextCNN & 0.7786  & 0.8011  & 0.9040  & 0.8079  & 0.8692  & 0.9023  \\
          & RoBERTa & 0.7810  & 0.8583  & 0.9288  & 0.8184  & 0.8802  & 0.9108  \\
    \midrule
    \multicolumn{1}{c}{\multirow{5}[2]{*}{\begin{sideways}mixed\end{sideways}}} & BiGRU & 0.7479  & 0.7339  & 0.7448  & 0.7501  & 0.8321  & 0.8504  \\
          & TextCNN & 0.7519  & 0.7040  & 0.8322  & 0.7679  & 0.8362  & 0.8674  \\
          & RoBERTa & 0.7823  & 0.7967  & 0.9014  & 0.8101  & 0.8744  & 0.9058  \\
          & StyleLSTM & 0.8007  & 0.7937  & 0.9252  & 0.8285  & 0.8826  & 0.9250  \\
          & DualEmo & 0.8056  & 0.7868  & 0.9019  & 0.8270  & 0.8818  & 0.9251  \\
    \midrule
    \multicolumn{1}{c}{\multirow{5}[4]{*}{\begin{sideways}multi\end{sideways}}} & EANN  & 0.7937  & 0.7558  & 0.8836  & 0.8123  & 0.8743  & 0.9053  \\
          & MMoE  & 0.8022  & 0.8477  & \underline{0.9379}  & 0.8361  & 0.8920  & \underline{0.9265}  \\
          & MoSE  & 0.7981  & \textbf{0.8576 } & 0.9326  & 0.8318  & 0.8885  & 0.9252  \\
          & EDDFN & 0.8067  & \underline{0.8505}  & 0.9306  & 0.8378  & 0.8912  & 0.9263  \\
          & MDFEND & \underline{0.8080}  & 0.8473  & 0.9331  & \underline{0.8390}  & \underline{0.8936}  & 0.9237  \\
\cmidrule{2-8}          & M$^3$FEND & \textbf{0.8237** } & 0.8478  & \textbf{0.9392} & \textbf{0.8517**} & \textbf{0.8977*} & \textbf{0.9342*} \\
    \bottomrule
    \end{tabular}\label{tab:en3}
\end{minipage}%
\begin{minipage}{.51\linewidth}
\setlength{\tabcolsep}{1.5pt}
\centering
\caption{Results on the Ch-3 dataset.  * ($p \le 0.05$) and ** ($p \le 0.005$) indicate paired t-test of M$^3$FEND vs. the best baseline.}
    \begin{tabular}{cc||ccc||ccc}
    \toprule
    \multirow{2}[1]{*}{} & \multirow{2}[1]{*}{Method} & \multirow{2}[1]{*}{Politics} & \multirow{2}[1]{*}{Health} &  \multirow{2}[1]{*}{Ent.} & \multicolumn{3}{c}{overall} \\
    \cline{6-8} &&&& & F1    & Acc   & AUC \\
    \midrule
    \multicolumn{1}{c}{\multirow{3}[2]{*}{\begin{sideways}single\end{sideways}}} & BiGRU & 0.8469  & 0.8335  & 0.7913  & 0.8402  & 0.8411  & 0.9213  \\
          & TextCNN & 0.8514  & 0.9041  & 0.8423  & 0.8846  & 0.8850  & 0.9521  \\
          & RoBERTa & 0.8137  & 0.8924  & 0.8434  & 0.8691  & 0.8697  & 0.9420  \\
    \midrule
    \multicolumn{1}{c}{\multirow{5}[2]{*}{\begin{sideways}mixed\end{sideways}}} & BiGRU & 0.8384  & 0.8577  & 0.8687  & 0.8733  & 0.8741  & 0.9402  \\
          & TextCNN & 0.8579  & 0.8716  & 0.8683  & 0.8833  & 0.8838  & 0.9493  \\
          & RoBERTa & 0.8300  & 0.8955  & 0.8862  & 0.8911  & 0.8915  & 0.9566  \\
          & StyleLSTM & 0.8298  & 0.8924  & 0.8896  & 0.8912  & 0.8917  & 0.9564  \\
          & DualEmo & 0.8362  & 0.8968  & 0.9020  & 0.8977  & 0.8980  & 0.9605  \\
    \midrule
    \multicolumn{1}{c}{\multirow{5}[4]{*}{\begin{sideways}multi\end{sideways}}} & EANN  & 0.8405  & 0.9189  & 0.8974  & 0.9038  & 0.9042  & 0.9644  \\
          & MMoE  & \textbf{0.8779 } & 0.9215  & 0.8800  & 0.9048  & 0.9052  & 0.9629  \\
          & MoSE  & 0.8564  & 0.9023  & 0.8872  & 0.8978  & 0.8985  & 0.9572  \\
          & EDDFN & 0.8440  & 0.9235  & 0.8748  & 0.8965  & 0.8970  & 0.9614  \\
          & MDFEND & 0.8555  & \underline{0.9419}  & \underline{0.9103}  & \underline{0.9205}  & \underline{0.9208}  & \underline{0.9750}  \\
\cmidrule{2-8}          & M$^3$FEND & \underline{0.8618}  & \textbf{0.9479* } & \textbf{0.9304**} & \textbf{0.9308**} & \textbf{0.9311**} & \textbf{0.9759} \\
    \bottomrule
    \end{tabular}\label{tab:ch3}
\end{minipage}
\end{table*}

\begin{table*}[t]
\setlength{\abovecaptionskip}{3pt}
\setlength{\belowcaptionskip}{5pt}
  \centering
  \caption{Results on the Ch-6 dataset.  * ($p \le 0.05$) and ** ($p \le 0.005$) indicate paired t-test of M$^3$FEND vs. the best baseline.}
    \begin{tabular}{cc||cccccc||ccc}
    \toprule
    \multirow{2}[4]{*}{} & \multirow{2}[1]{*}{Method} & \multirow{2}[1]{*}{Edu.} & \multirow{2}[1]{*}{Disaster} & \multirow{2}[1]{*}{Health} & \multirow{2}[1]{*}{Finance} & \multirow{2}[1]{*}{Ent.} & \multirow{2}[1]{*}{Society} & \multicolumn{3}{c}{overall} \\
\cline{9-11}          &       &       &       &       &       &       &       & F1    & Acc   & AUC \\
    \midrule
    \multicolumn{1}{c}{\multirow{3}[2]{*}{\begin{sideways}single\end{sideways}}} & BiGRU & 0.7697  & 0.7191  & 0.8451  & 0.8247  & 0.8026  & 0.8015  & 0.8266  & 0.8270  & 0.8979  \\
          & TextCNN & 0.7805  & 0.4388  & 0.9012  & 0.7671  & 0.7930  & 0.8654  & 0.8494  & 0.8499  & 0.9195  \\
          & RoBERTa & 0.8175  & 0.7584  & 0.8909  & 0.8498  & 0.8549  & 0.8304  & 0.8576  & 0.8580  & 0.9288  \\
    \midrule
    \multicolumn{1}{c}{\multirow{5}[1]{*}{\begin{sideways}mixed\end{sideways}}} & BiGRU & 0.8253  & 0.7938  & 0.8626  & 0.8254  & 0.8604  & 0.8206  & 0.8491  & 0.8501  & 0.9249  \\
          & TextCNN & 0.8593  & 0.8240  & 0.8832  & 0.8646  & 0.8659  & 0.8641  & 0.8776  & 0.8783  & 0.9483  \\
          & RoBERTa & 0.8664  & 0.8515  & 0.9100  & 0.8700  & 0.8872  & 0.8634  & 0.8872  & 0.8877  & 0.9494  \\
          & StyleLSTM & 0.8565 & 0.8374 & 0.9080 & 0.8766 & 0.8957 & 0.8546 & 0.8844 & 0.8851 & 0.9489\\
          & DualEmo & 0.8472  & 0.8352  & 0.9055  & \underline{0.8951}  & {0.9043}  & 0.8642  & 0.8904  & 0.8909  & 0.9579  \\
    \midrule
    \multicolumn{1}{c}{\multirow{5}[3]{*}{\begin{sideways}multi\end{sideways}}} & EANN  & 0.8613  & 0.8657  & 0.9150  & 0.8621  & 0.8871  & {0.8791}  & {0.8919}  & {0.8925}  & {0.9605}  \\
          & MMoE  & 0.8625  & {0.8777}  & 0.9260  & 0.8546  & 0.8882  & 0.8655  & 0.8894  & 0.8900  & 0.9563  \\
          & MoSE  & 0.8569  & 0.8588  & 0.9118  & 0.8639  & 0.8904  & 0.8757  & 0.8913  & 0.8918  & 0.9533  \\
          & EDDFN & {0.8780}  & 0.8734  & {0.9280}  & 0.8456  & 0.8819  & 0.8716  & 0.8917  & 0.8921  & 0.9544  \\
          & MDFEND & \underline{0.8826}  & \underline{0.8781}  & \underline{0.9430}  & {0.8749}  & \underline{0.9095}  & \underline{0.8940}  & \underline{0.9093}  & \underline{0.9097}  & \underline{0.9694}  \\
\cmidrule{2-11}          & M$^3$FEND & \textbf{0.8836 } & \textbf{0.8824 } & \textbf{0.9515*} & \textbf{0.8997*} & \textbf{0.9296**} & \textbf{0.9043**} & \textbf{0.9208**} & \textbf{0.9211**} & \textbf{0.9762*} \\
    \bottomrule
    \end{tabular}%
  \label{tab:ch6}%
\end{table*}%

To capture multi-view information, we extend the two datasets with emotion and style features. All emotion features are listed in Table~\ref{tab:emofea}. The extraction process of emotion features follows the work of Zhang et al.~\cite{zhang2021mining} with their official codes.\footnote{https://github.com/RMSnow/WWW2021} In this paper, we follow Yang et al.~\cite{yang2019write} to extract Style features, including eight high-level features:
\begin{itemize}[leftmargin=15pt]
    \item \textit{Readability} measures the   clarity of the news, evaluated as: Readability = - (Sentence\_broken + Characters + Words + Sentences + Clauses + Average\_word length + Professional\_words+ LW + RIX + LIX).
    \item \textit{Logic} determines whether the news  are logical and contextually coherent or not. Logic = Forward\_reference + Conjs.
    \item \textit{Credibility} measures the rigor and reliability of the news, computed as: Credibility = @ + Numerals + Official speech + Time + Place + Object - Uncertainty + Image.
    \item \textit{Formality} is used to measure the writing normative. Formality = Noun + Adj + Prep - Pron - Verb - Adv - Sentence\_broken.
    \item \textit{Interactivity} represents the interaction between news and the readers, computed as: Interactivity = Question\_mark + First\_pron + Second\_pron + Interrogative\_pron.
    \item \textit{Interestingness} is used to measure whether the news will be attractive. Interestingness = Rhetoric + Exclamation mark + Face + Idiom + Adversative + Adj + Image.
    \item \textit{Sensation} measures the impression that the news leaves on the reader, computed as: Sensation = Sentiment\_score + Adv\_of\_degree + Modal\_particle + First\_pron + Second\_pron + Exclamation\_mark + Question\_mark.
    \item \textit{Integrity} is used to measure whether a news piece is complete. Integrity = 2*HasHead + 2*HasImage + 2*HasVideo + 2*HasTag + HasAt + HasUrl.
\end{itemize}

We just list how to calculate these style features based on other low-level features, and more details can be found in~\cite{yang2019write}. Both low- and high-level style features are exploited to extract information of the stylistic view in this paper.

\subsubsection{Baselines.} 
We categorize our baselines into three groups. The first group is single-domain methods that separately train models for each domain, including:
\begin{itemize}[leftmargin=15pt]
    \item BiGRU~\cite{ma2016detecting} is a widely used baseline in many existing works of fake news detection for text encoding. We implement a one-layer BiGRU with a hidden size of 300.
    \item TextCNN~\cite{kim-2014-convolutional} is a popular text encoder. We implement TextCNN with 5 kernels. The 5 kernels with the same 64 channels have different steps of 1, 2, 3, 5, and 10.
    \item RoBERTa~\cite{liu2019roberta,cui2019pre} is a robustly optimized BERT~\cite{devlin2019bert} pre-training model. We utilize RoBERTa to encode tokens of news content and feed the extracted average embedding into an MLP to obtain the final prediction. The two kinds of RoBERTa~\cite{liu2019roberta,cui2019pre} are exploited for Chinese and English datasets, respectively.
\end{itemize}

\begin{table*}[t]
\setlength{\abovecaptionskip}{3pt}
\setlength{\belowcaptionskip}{5pt}
  \centering
  \setlength{\tabcolsep}{2.5pt}
  \caption{Results on the Ch-9 dataset. * ($p \le 0.05$) and ** ($p \le 0.005$) indicate paired t-test of M$^3$FEND vs. the best baseline.}
    \begin{tabular}{cc||ccccccccc||ccc}
    \toprule
    \multirow{2}[4]{*}{} & \multirow{2}[1]{*}{Method} & \multirow{2}[1]{*}{Science} & \multirow{2}[1]{*}{Military} & \multirow{2}[1]{*}{Edu.} & \multirow{2}[1]{*}{Disaster} & \multirow{2}[1]{*}{Politics} & \multirow{2}[1]{*}{Health} & \multirow{2}[1]{*}{Finance} & \multirow{2}[1]{*}{Ent.} & \multirow{2}[1]{*}{Society} & \multicolumn{3}{c}{overall} \\
\cline{12-14}          &       &       &       &       &       &       &       &       &       &       & F1    & Acc   & AUC \\
    \midrule
    \multicolumn{1}{c}{\multirow{3}[1]{*}{\begin{sideways}single\end{sideways}}} & BiGRU & 0.5175  & 0.3365  & 0.7416  & 0.7293  & 0.8588  & 0.8373  & 0.8137  & 0.7992  & 0.7918  & 0.8103  & 0.8103  & 0.8902  \\
          & TextCNN & 0.4074  & 0.3365  & 0.8059  & 0.4388  & 0.8482  & 0.8819  & 0.8215  & 0.7973  & 0.8615  & 0.8369  & 0.8370  & 0.9094  \\
          & RoBERTa & 0.7463  & 0.7369  & 0.8146  & 0.7547  & 0.8044  & 0.8873  & 0.8361  & 0.8513  & 0.8300  & 0.8477  & 0.8477  & 0.9226  \\
    \midrule
    \multicolumn{1}{c}{\multirow{5}[1]{*}{\begin{sideways}mixed\end{sideways}}} & BiGRU & 0.7269  & 0.8724  & 0.8138  & 0.7935  & 0.8356  & 0.8868  & 0.8291  & 0.8629  & 0.8485  & 0.8595  & 0.8598  & 0.9309  \\
          & TextCNN & 0.7254  & 0.8839  & 0.8362  & 0.8222  & 0.8561  & 0.8768  & 0.8638  & 0.8456  & 0.8540  & 0.8686  & 0.8687  & 0.9381  \\
          & RoBERTa & 0.7777  & 0.9072  & 0.8331  & 0.8512  & 0.8366  & 0.9090  & 0.8735  & 0.8769  & 0.8577  & 0.8795  & 0.8797  & 0.9451  \\
          & StyleLSTM & 0.7729  & 0.9187  & 0.8341  & 0.8532  & 0.8487  & 0.9084  & 0.8802  & 0.8846  & 0.8552  & 0.8820  & 0.8821  & 0.9471  \\
            & DualEmo & 0.8323  & 0.9026  & 0.8362  & 0.8396  & 0.8455  & 0.8905  & \textbf{0.9053}  & 0.8944  & 0.8569  & 0.8846 &  0.8846 & 0.9541 \\
    \midrule
    \multicolumn{1}{c}{\multirow{5}[4]{*}{\begin{sideways}multi\end{sideways}}} & EANN  & 0.8225  & {0.9274}  & 0.8624  & 0.8666  & 0.8705  & 0.9150  & 0.8710  & {0.8957}  & {0.8877}  & {0.8975}  & {0.8977}  & {0.9610}  \\
          & MMoE  & \textbf{0.8755 } & 0.9112  & 0.8706  & 0.8770  & 0.8620  & 0.9364  & 0.8567  & 0.8886  & 0.8750  & 0.8947  & 0.8948  & 0.9547  \\
          & MoSE  & \underline{0.8502}  & 0.8858  & {0.8815}  & 0.8672  & {0.8808}  & 0.9179  & 0.8672  & 0.8913  & 0.8729  & 0.8939  & 0.8940  & 0.9543  \\
          & EDDFN & 0.8186  & 0.9137  & 0.8676  & {0.8786}  & 0.8478  & {0.9379}  & 0.8636  & 0.8832  & 0.8689  & 0.8919  & 0.8919  & 0.9528  \\
          & MDFEND & 0.8301  & \underline{0.9389}  & \underline{0.8917}  & \textbf{0.9003 } & \textbf{0.8865}  & \underline{0.9400}  & {0.8951}  & \underline{0.9066}  & \underline{0.8980}  & \underline{0.9137}  & \underline{0.9138}  & \underline{0.9708}  \\
\cmidrule{2-14}          & M$^3$FEND & 0.8292  & \textbf{0.9506** } & \textbf{0.8998} & \underline{0.8896}  & \underline{0.8825}  & \textbf{0.9460} & \underline{0.9009 } & \textbf{0.9315**} & \textbf{0.9089**} & \textbf{0.9216**} & \textbf{0.9216**} & \textbf{0.9750*} \\
    \bottomrule
    \end{tabular}%
  \label{tab:ch9}%
\end{table*}%

The second group is the mixed-domain baselines which combine all domains into a single domain.  BiGRU~\cite{ma2016detecting}, TextCNN~\cite{kim-2014-convolutional}, and RoBERTa~\cite{liu2019roberta,cui2019pre} of this group have the same implementation as the first group. Another two baselines of this groups are:
\begin{itemize}[leftmargin=15pt]
    \item StyleLSTM~\cite{przybyla2020capturing} exploits a BiLSTM to extract news representation from content. Then, it feeds the representation and style features into an MLP to obtain the final prediction. The utilized style features in this paper are the same as StyleLSTM.
    \item DualEmo~\cite{zhang2021mining} extracts news representation with a BiGRU. It exploits both the representation and emotion features to predict fake or real. The utilized emotion features in this paper are the same as DualEmo.
\end{itemize}

The third group is well-designed multi-domain methods, including:
\begin{itemize}[leftmargin=15pt]
    \item EANN~\cite{wang2018eann} consists of three components: feature extractor, event discriminator, and fake news detector. It aims to learn event-invariant representations. In this paper, we modify EANN to learn domain-invariant representations following~\cite{silva2021embracing}.
    \item MMoE~\cite{ma2018modeling} is a popular multi-domain model that shares a mixture-of-experts (MoE) across various domains, and each domain has its specific head. In this paper, both experts and heads of MMoE are MLPs.
    \item MoSE~\cite{qin2020multitask} is a recent multi-domain model that replaces the experts of MMoE with LSTM.
    \item EDDFN~\cite{silva2021embracing} is a multi-domain fake news detection model which preserves domain-specific and domain-shared knowledge. All domain-specific and domain-shared modules are MLPs. The concentrated domain-specific and domain-shared representations are fed into a classifier to obtain the final prediction.
    \item MDFEND~\cite{nan2021mdfend} is the latest multi-domain fake new detection model which utilizes a Domain Gate to select useful experts of MoE.
\end{itemize}



\subsubsection{Experimental details.} Following~\cite{shu2019defend,nan2021mdfend}, all datasets are randomly divided into train / validation / test sets in the ratio of 6:2:2 with keep the domain distribution in each set. We do not perform any dataset-specific tuning except early stopping on validation sets. 
For all methods, the initial learning rate for the Adam~\cite{kingma2014adam} optimizer is tuned by grid searches from 1e-6 to 1e-2. The number of heads and channels is tuned by grid searches from 1 to 10. The number of memory units $Q$ is searched from 5 to 50. The mini-batch size is 64. To ensure a fair comparison, all BiGRU and BiLSTM have one layer with a hidden size of 300. All TextCNN have 5 kernels with steps in \{1, 2, 3, 5, 10\} and 64 channels. For all MLPs in these methods, the dimension of the hidden layer is set to 384, and ReLU activation function is employed. The maximum sequence length of English and Chinese datasets is set as 300 and 170, respectively. We record average results over ten runs. We report accuracy (Acc), macro F1 score (F1), and Area Under ROC (AUC).

\begin{table}[t]
\setlength{\abovecaptionskip}{3pt}
\setlength{\belowcaptionskip}{5pt}
  \centering
  \caption{Relative improvement over the online baseline.}
    \begin{tabular}{cccc}
    \toprule
    \multicolumn{1}{l}{Improvement on} & SPAUC & AUC   & F1 \\
    \midrule
    EANN  & 2.12\% & 0.67\% & 0.33\% \\
    EDDFN & -0.37\% & -2.02\% & -3.34\% \\
    MDFEND & 2.82\% & 0.74\% & 1.85\% \\
    M$^3$FEND & \textbf{5.50\%} & \textbf{2.89\%} & \textbf{4.49\%} \\
    \bottomrule
    \end{tabular}%
  \label{tab:online}%
\end{table}%

\subsection{Offline Results (RQ1)}\label{sec:offline_experiment}
In this section, we conduct offline experiments on both English and Chinese datasets. The results of En-3 are shown in Table~\ref{tab:en3}, and results of Ch-3, -6, and -9 are listed in Tables~\ref{tab:ch3}, ~\ref{tab:ch6}, and ~\ref{tab:ch9} (F1 for each domain and F1, Acc, and AUC for overall performance), respectively. Bold and underlined results indicate the best and second best, respectively. From these results, we have several findings:

(1) On most tasks, the mixed-domain methods outperform the single-domain approaches, which demonstrates jointly training data of multiple domains is helpful to improve not only the overall performance of multiple domains but also the performance of each domain. Meanwhile, we observe that BiGRU, TextCNN, and RoBERTa of the second group have worse performance than the single-domain group in Table~\ref{tab:en3}, and this phenomenon could be caused by serious domain conflict of the En-3 dataset, which shows the necessity of well-designed multi-domain models to alleviate domain conflict. 

(2) Then, we find that the multi-domain group outperforms the mixed-domain group, which validates that well-designed multi-domain models are important. The reason could be that the multi-domain models with a well-designed sharing structure could alleviate the domain conflict.

(3) StyleLSTM and DualEmo achieve better results than BiGRU, TextCNN, and RoBERTa, which testifies introducing more views is beneficial for multi-domain fake news detection. Especially, DualEmo is extremely effective for Entertainment and Finance domains and even outperforms most multi-domain methods.

(4) With the results of the t-test, we find that M$^3$FEND outperforms the best baseline significantly in most tasks, which demonstrates that M$^3$FEND is an effective solution to improve not only overall detection performance but also the performance of the specific domains. The main reason is that M$^3$FEND enriches domain information and explicitly models various domain discrepancies by aggregating useful cross-view interactions for different domains.

(5) We observe MDFEND, and M$^3$FEND outperforms EANN, MMoE, MoSE, and EDDFN on most tasks. EANN directly learns a shared network with adversarial training. MMoE and MoSE utilized a shared bottom and multiple separate heads for different domains. EDDFN learns both domain-specific and domain-shared subnetworks. We find that all of them take hard sharing mechanisms to learn shared knowledge from all domains. However, the existing studies~\cite{zhu2019aligning,tang2020progressive} found it is hard to learn the common knowledge with a shared structure from too many domains. Different from these methods, MDFEND and M$^3$FEND exploit soft sharing mechanisms to aggregate beneficial shared knowledge for fake news detection.

(6) Both MDFEND and M$^3$FEND exploit soft sharing mechanisms, but M$^3$FEND achieves better results on most tasks. The improvements come from multiple aspects: 1) M$^3$FEND contains a Domain Memory Bank to discover potential distributions of domain labels, while MDFEND assumes the given domain label is exact and complete. However, we find that domain labeling incompleteness is an important issue for multi-domain fake news detection in Section~\ref{sec:analysis}; 2) MDFEND only extracts semantic information, while M$^3$FEND simultaneously models semantic, emotional, and stylistic views and adaptively captures cross-view information.

\subsection{Online Tests (RQ2)}
The M$^3$FEND framework has already been deployed in our online fake news detection system (http://www.newsverify.com/) which handles millions of news pieces every day. 
To verify the real benefits of M$^3$FEND brings to our system, we conduct online testing experiments within one week. Different from offline datasets, the online test set is highly skewed (real vs. fake, roughly 300:1). Online data is collected by the system, and the time intervals of training and test sets do not intersect. Precisely, we deploy M$^3$FEND and several competitive baselines (EANN, EDDFN, and MDFEND). The online baseline is the mixed-domain RoBERTa model. In real-world scenarios, the number of fake news is much lower than real news, which means that we should detect fake news without misclassifying real news as possible. In other words, the task is improving the True Positive Rate (TPR) on the basis of low False Positive Rate (FPR). Thus, beyond AUC and F1, following~\cite{mcclish1989analyzing,zhu2020modeling}, we adopt standardized partial AUC (SPAUC$_{\text{FPR}\leq 0.1}$). Due to the company regulations, we cannot detail the online data and the absolute results, so we report the relative improvement over the online baseline RoBERTa. The online results in Table~\ref{tab:online} demonstrate that the proposed M$^3$FEND achieves a satisfying improvement on all metrics against the baselines.

\begin{table}[t]
\setlength{\abovecaptionskip}{3pt}
\setlength{\belowcaptionskip}{-3pt}
  \centering
  \caption{Results of ablation study.}
    \begin{tabular}{ccccc}
    \toprule
          & Ch-3 & Ch-6 & Ch-9 & En-3 \\
    \midrule
    M$^3$FEND & \textbf{0.9308} & \textbf{0.9208} & \textbf{0.9216} & \textbf{0.8517} \\
    \midrule
    w/o SemView & 0.8202 & 0.8161 & 0.8249 & 0.6573 \\
    w/o EmoView & 0.9195 & 0.9136 & 0.9147 & 0.8403 \\
    w/o StyView & 0.9255 & 0.9178 & 0.9177 & 0.8472 \\
    \midrule
    w/o Interactor & 0.9217 & 0.9169 & 0.9173 & 0.8398 \\
    w/o Memory & 0.9237 & 0.9182 & 0.9176 & 0.8501 \\
    w/o Adapter & 0.9172 & 0.9169 & 0.9157 & 0.8367 \\
    \bottomrule
    \end{tabular}%
  \label{tab:ablation}
\end{table}%

\begin{table*}[t]
\setlength{\abovecaptionskip}{3pt}
\setlength{\belowcaptionskip}{5pt}
  \centering
  \caption{A case of the distribution of predicted domain label.}
    \begin{tabular}{ccl}
    \toprule
    \multicolumn{2}{c}{Target News} & Trump nearly fainted during his speech and cancelled his subsequent trip. A symptom of COVID-19? \\
    \midrule
    \midrule
    Domain & Similarity $\bm{v}$ & Representative Example \\
    \midrule
    Science & \cellcolor[rgb]{ .847,  .906,  .957}0.02 & NASA used the Nuclear Spectroscopy Telescope to photo the spiral galaxy 1068 in the Cetus. \\
    Military & \cellcolor[rgb]{ .8,  .867,  .925}0.04 & U.S. sends 35 medical ships. \\
    Edu.  & \cellcolor[rgb]{ .867,  .922,  .969}0.01 & A student admitted to Harvard University. \\
    Disaster & \cellcolor[rgb]{ .847,  .906,  .957}0.02 & The US "World Journal" reported a five-level fire in a restaurant. \\
    Politics & \cellcolor[rgb]{ .122,  .306,  .471}0.33 & US deaths from COVID-19 exceed 100k. \\
    Health & \cellcolor[rgb]{ .404,  .537,  .659}0.21 & The animal experiment of Oxford's COVID-19 vaccine failed. \\
    Finance & \cellcolor[rgb]{ .612,  .714,  .8}0.12 & Pfizer’s stocking price rose 15\%, boosted by the company's COVID-19 vaccine news. \\
    Ent.  & \cellcolor[rgb]{ .682,  .769,  .847}0.09 & 10 more people tested positive for COVID-19 in Italian Serie A. \\
    Society & \cellcolor[rgb]{ .518,  .635,  .737}0.16 & A COVID-19 carrier refused security check at the airport. \\
    \bottomrule
    \end{tabular}%
  \label{tab:case}%
\end{table*}%

\subsection{Ablation Study (RQ3)}
In this section, we analyze the effects of different views and components in our proposed M$^3$FEND and conduct an ablation study on the four datasets with the overall F1 score shown in Table~\ref{tab:ablation}. First, we conduct experiments to verify the contributions of different views and introduce three kinds of models, w/o SemView, w/o EmoView, and w/o StyView, which remove the semantic, emotional, and stylistic views from M$^3$FEND, respectively. We find that all views are beneficial for fake news detection, especially the semantic view, which is the core of most existing methods~\cite{ma2019detect,shu2019defend}. Since the emotion and style features are manually extracted from the textual content, these features are usually utilized as auxiliary information for semantic view modeling~\cite{przybyla2020capturing,zhang2021mining}. In addition, we observe that the emotional view is more effective than the stylistic view. The reason could be that the emotion features include both publisher and social characteristics~\cite{zhang2021mining}, while the style features only represent the publisher preference~\cite{yang2019write}.

Furthermore, to testify the effectiveness of each component in M$^3$FEND, we introduce three kinds of M$^3$FEND, (1) w/o Interactor: remove Multi-head Adaptive Cross-view Interactor. The performance drop demonstrates modeling cross-view interaction could capture more information. (2) w/o Memory: remove the Domain Event Memory. The performance decrease on four datasets indicates enriching domain information is useful. (3) w/o Adapter: replace Domain Adapter by average operation. M$^3$FEND w/o Adapter obtains the worse results, which demonstrates the aggregation process of discriminative representations is necessary. Note that the Adapter takes two parts from the Domain Event Memory and Domain Characteristics Memory as input. Thus, that M$^3$FEND w/o Adapter achieves worse results than M$^3$FEND w/o Memory demonstrates the implicit representation from the Domain Characteristics Memory is useful. In addition, we observe the Adapter is more effective than the Interactor, which shows that selecting informative representations is more effective than adaptively learning cross-view information.

\begin{figure}[t]
\setlength{\abovecaptionskip}{3pt}
\setlength{\belowcaptionskip}{-10pt}
\centering
\begin{minipage}[b]{1\linewidth}
\centering
\includegraphics[width=1\linewidth]{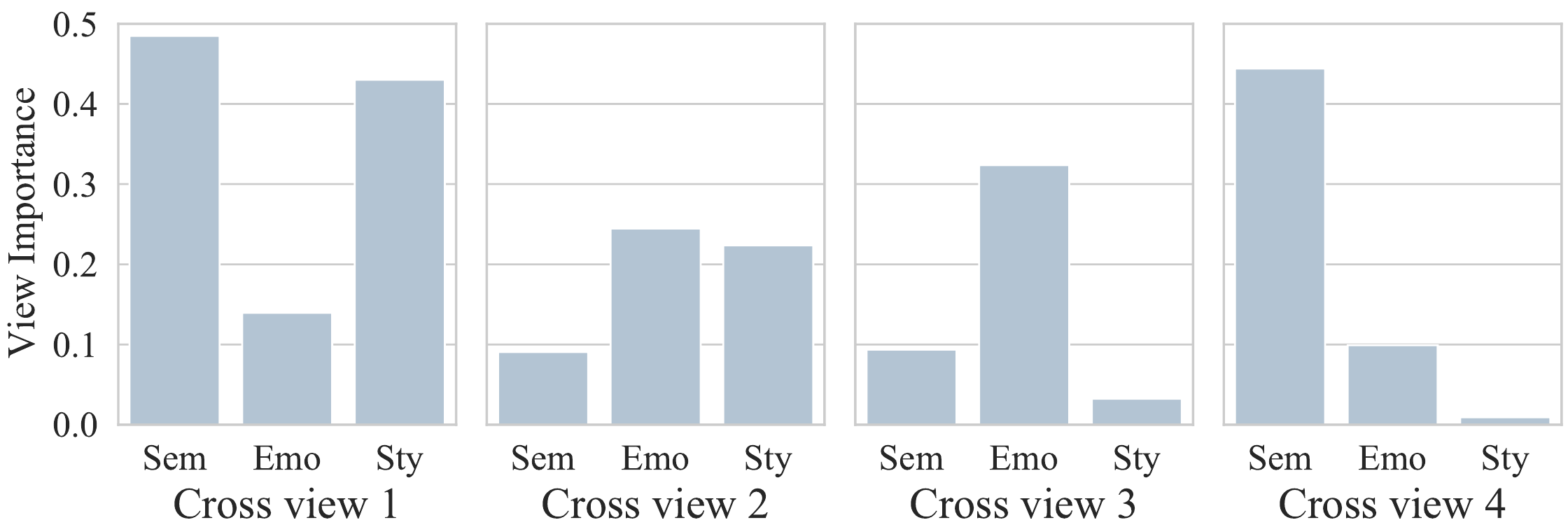}
\end{minipage}
\caption{Each figure indicates importances of different views in a cross-view interaction.}\label{fig:cross_experiment1}
\end{figure} 

\begin{figure}[t]
\setlength{\abovecaptionskip}{3pt}
\setlength{\belowcaptionskip}{-10pt}
\centering
\begin{minipage}[b]{1\linewidth}
\centering
\includegraphics[width=1\linewidth]{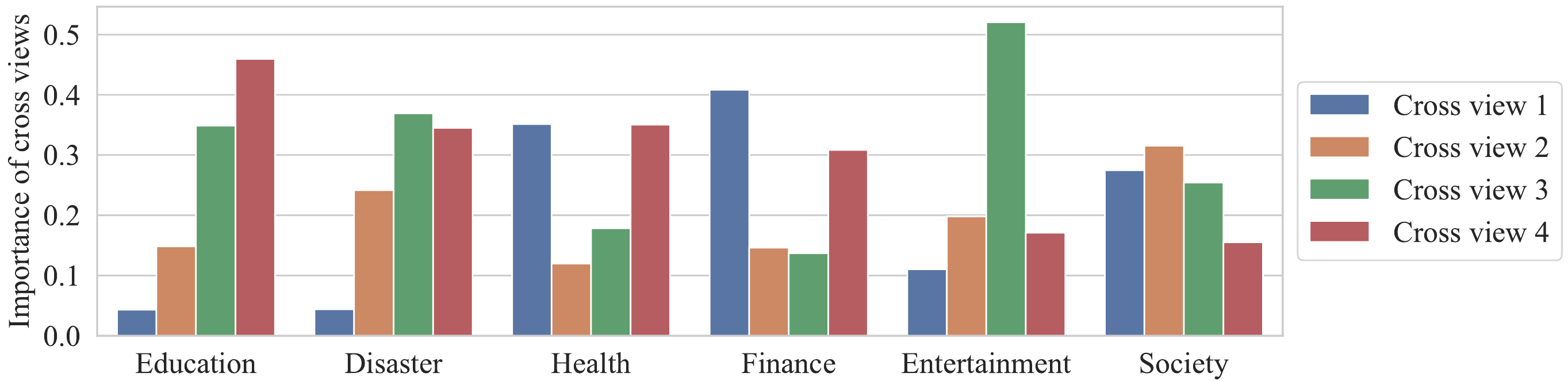}
\end{minipage}
\caption{Various importances of four cross-view interactions for different domains.   }\label{fig:cross_experiment2}
\end{figure}

\subsection{Analysis (RQ4)}
\subsubsection{Effectiveness of Domain Discrepancy Modeling}
In this section, we conduct experiments on Ch-6 to demonstrate M$^3$FEND can model the domain discrepancy. Due to domain discrepancy, discriminative cross-view representations vary from domain to domain, and the proposed M$^3$FEND can find useful cross-view representations for different domains. For brevity, we set the number $H$ of Interactor heads as 4 and the channel number $k$ of multi-view extractors as 1, and remove the implicit domain representations. Firstly, we visualize the absolute adaptive weight $a$ in Equation~\ref{eq:cross_feature} for four cross views, as shown in Figure~\ref{fig:cross_experiment1}, we see that the cross-view representations derived from M$^3$FEND have quite differences on the combination of semantic, emotional, and stylistic views, which diversifies cross-view representations and facilitates the modeling of domain discrepancy. 

Then, the domain adapter aggregates useful cross-view representations for each domain, and we visualize the importance $\bm{w}$ of each cross-view representation for six domains in Figure~\ref{fig:cross_experiment2}. We see the discriminative cross-view representations vary from domain to domain, which demonstrates the M$^3$FEND can model the domain discrepancy. Such importance distributions are indeed in line with our intuition. For example, the Entertainment news is often sensational, which provokes strong emotion of audience. We see the cross view 3 which largely depends on the emotional view is valued the most by Entertainment Domain.

\subsubsection{Effectiveness of Domain Label Completion}
To probe how the Domain Event Memory contributes to discovering potential domain labels, we present a case study. Recall that a unit of a Domain Event Memory represents an event set, and each news piece is only categorized into one unit. For the target political news piece in Table~\ref{tab:case}, we find the most related memory units of each Domain Event Memory and show a representative example from the event sets of them. The predicted distribution of domain labels (similarity distribution $\bm{v}$) indicates that the Domain Event Memory can effectively capture the distributions of potential domain labels.

\begin{figure*}[ht]
\setlength{\abovecaptionskip}{3pt}
\setlength{\belowcaptionskip}{-11pt}
\centering
\begin{minipage}[b]{1\linewidth}
\centering
\includegraphics[width=1\linewidth]{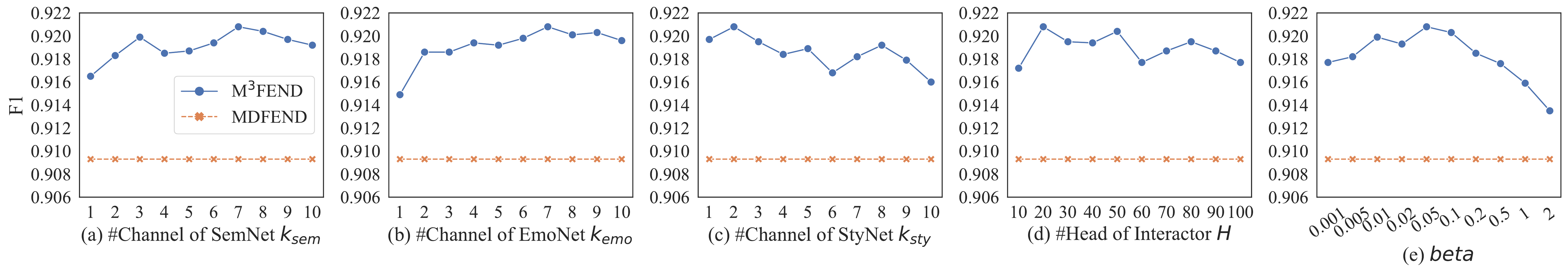}
\end{minipage}
\caption{Performance (F1) of M$^3$FEND with various hyperparameters.}\label{fig:hyperparameter}
\end{figure*}

\subsubsection{Hyperparameter Sensitivity} 
We test the sensitivity of multiple hyperparameters based on the Ch-6 dataset, including \#Channel of SemNet $k_{sem}$, \#Channel of EmoNet $k_{emo}$, \#Channel of StyNet $k_{sty}$, \#Head of Interactor $H$, and the hyperparameter of the memory mechanism $\beta$. As shown in Figure~\ref{fig:hyperparameter}, with various hyperparameters, M$^3$FEND can achieve satisfying performance. We can observe that even the worst setting of M$^3$FEND in Figure~\ref{fig:hyperparameter}(a)-(d) can get F1=0.9149 which is better than the best baseline MDFEND (F1:0.9093) on the Ch-6 dataset, which demonstrates our model are insensitive to the hyperparameters $k_{sem}$, $k_{emo}$, $k_{sty}$, and $H$.  From Figure~\ref{fig:hyperparameter}(e), we can observe that the performance first increases and then decreases rapidly as $\beta$ varies and demonstrates a bell-shaped curve. A big $\beta$ indicates that the memory module could quickly forget historical samples and focus on most recent samples, which leads to overfitting on recent samples. unsatisfying performance. On the contrary, a small $\beta$ could lead to underfitting on recent samples. Thus, we need to choose a suitable $\beta$ for the Domain Event Memory.


\section{Related Work}
In this section, we will introduce the related work on Fake News Detection and Multi-domain Learning.

\textbf{Fake News Detection.} Researchers have investigated fake news detection which aims at automatically classifying a news piece
as real or fake for a long time. Existing methods can be generally grouped into two clusters: content-based and social-context-based fake news detection~\cite{shu2017fake,zhou2020survey}. 

Content-based models mainly rely on news content features and existing factual sources to classify fake news~\cite{shu2017fake,zhou2020survey}. Some methods focus on extracting textual representations~\cite{ma2019detect,karimi2019learning,kim2021effective,zhu2022generalizing,sheng-acl22}. In addition, visual features of news have been shown to be an important indicator for fake news detection~\cite{wang2018eann,wang2021metafend,qipeng-mm}. As fake news publishers tend to use inflammatory and emotional expressions to draw reader's attention for a wide dissemination, style~\cite{castillo2011information,potthast2018stylometric,przybyla2020capturing} and emotion~\cite{giachanou2019leveraging,zhang2021mining,moral-emotion} are useful patterns for fake news detection. Some methods~\cite{popat2018declare,cui2020deterrent,vo2021hierarchical,sheng2021integrating} exploit existing factual sources to detect fake news.

Social-context-based models exploit relevant user social engagements to detect fake news~\cite{shu2017fake}. Propagation networks have been testified their effectiveness for fake news detection~\cite{liu2018early,nguyen2020fang,silva2021propagation2vec}. In addition, user profile~\cite{shu2018understanding,dou2021user} and crowd feedbacks~\cite{ma2018detect,shu2019defend} are also important patterns to detect fake news.

Our work mainly falls into the first group, which utilizes semantic, style, emotion features to detect fake news. In addition, most of existing works focus on a single specific domain, e.g., politics~\cite{potthast2018stylometric,shu2019defend}, health~\cite{li2020mm,cui2020deterrent,zhou2020recovery}. Our work focuses on multi-domain fake news detection~\cite{silva2021embracing,nan2021mdfend}. Silva et al.~\cite{silva2021embracing} proposed EDDFN that adopts multiple domain-specific subnetworks and a domain-shared subnetwork, and models common knowledge from \textit{all} domains with the domain-shared subnetwork. However, studies have shown that it is hard to simultaneously model the common knowledge of many domains~\cite{zhu2019aligning,tang2020progressive}. Thus, EDDFN is unlikely to work well for a real-world news platform that has a large and increasing domain set. Different from EDDFN which uses a hard sharing mechanism, our M$^3$FEND adopts a soft sharing mechanism that aggregates the specific shared knowledge of different domains. Nan et al.~\cite{nan2021mdfend} built a multi-domain fake news detection dataset containing news in nine domains and a simple baseline MDFEND which also takes the soft sharing strategy and utilizes a domain gate to aggregate multiple semantic representations extracted by a Mixture-of-Experts structure~\cite{jacobs1991adaptive,ma2018modeling}. However, MDFEND holds an assumption that the provided single domain label is \textit{complete}, which is not always true due to the complex semantic property of news pieces. To tackle the problem of domain labeling incompleteness, our proposed M$^3$FEND utilizes the Domain Memory Bank to complete domain labels and enrich domain information in news pieces. In addition, we explore the mechanism of knowledge sharing on not only the semantic view but also emotional and stylistic views, and we further explore the adaptive cross-view knowledge for fake news detection.


\textbf{Multi-domain Learning.} In the real world, data usually come from various domains. Multi-domain learning aims to simultaneously model multiple domains and improve overall performance, which falls into the area of transfer learning~\cite{pan2009survey}. Some existing methods focus on learning domain-invariant representations~\cite{ganin2016domain,zhu2019aligning,peng2019moment}, and the others model domain relationships~\cite{ma2018modeling,qin2020multitask}. M$^3$FEND falls into the second group and outperforms existing methods.

\section{Conclusion}
In this paper, we analyzed two challenges in multi-domain fake news detection, domain shift, and domain labeling incompleteness. Then, to solve the two challenges, we proposed a novel Memory-guided Multi-view Multi-domain Fake News Detection Framework (M$^3$FEND). Firstly, we extracted news representations from multiple views and automatically modeled cross-view interactions. To tackle the problem of domain shift, we proposed a Domain Adapter to aggregate cross-view representations for prediction. To solve the second challenge, we proposed a Domain Memory Bank to discover potential domain labels and model domain characteristics. Finally, we demonstrated the superior effectiveness of the proposed M$^3$FEND in both offline experiments and online tests, compared to other state-of-the-art approaches.


%

\ifCLASSOPTIONcompsoc
  \section*{Acknowledgments}
\else
  \section*{Acknowledgment}
\fi

The research work is supported by the National Key Research and Development Program of China (2021AAA0140203), the Zhejiang Provincial Key Research and Development Program of China (2021C01164), the Project of Chinese Academy of Sciences (E141020), and the National Natural Science Foundation of China (62176014).

\ifCLASSOPTIONcaptionsoff
  \newpage
\fi

\bibliographystyle{IEEEtran}
\bibliography{reference}



%

%

\begin{IEEEbiography}[{\includegraphics[width=1in,height=1.25in,clip,keepaspectratio]{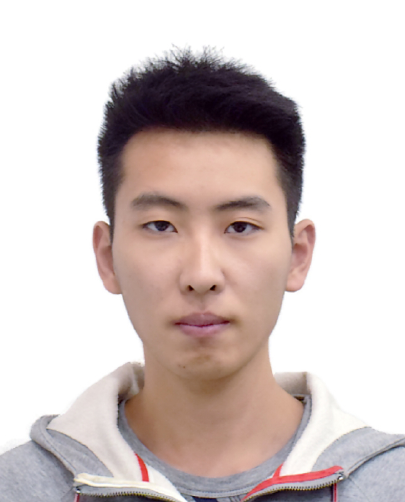}}]{Yongchun Zhu}
is currently pursuing his Ph.D. degree in the Institute of Computing Technology, Chinese Academy of Sciences, Beijing, China. His main research interests include transfer learning, fake news detection and recommender system. He has published over 20 papers in journals and conference proceedings including KDD, WWW, SIGIR, TNNLS and so on.
\end{IEEEbiography}

\begin{IEEEbiography}[{\includegraphics[width=1in,height=1.25in,clip,keepaspectratio]{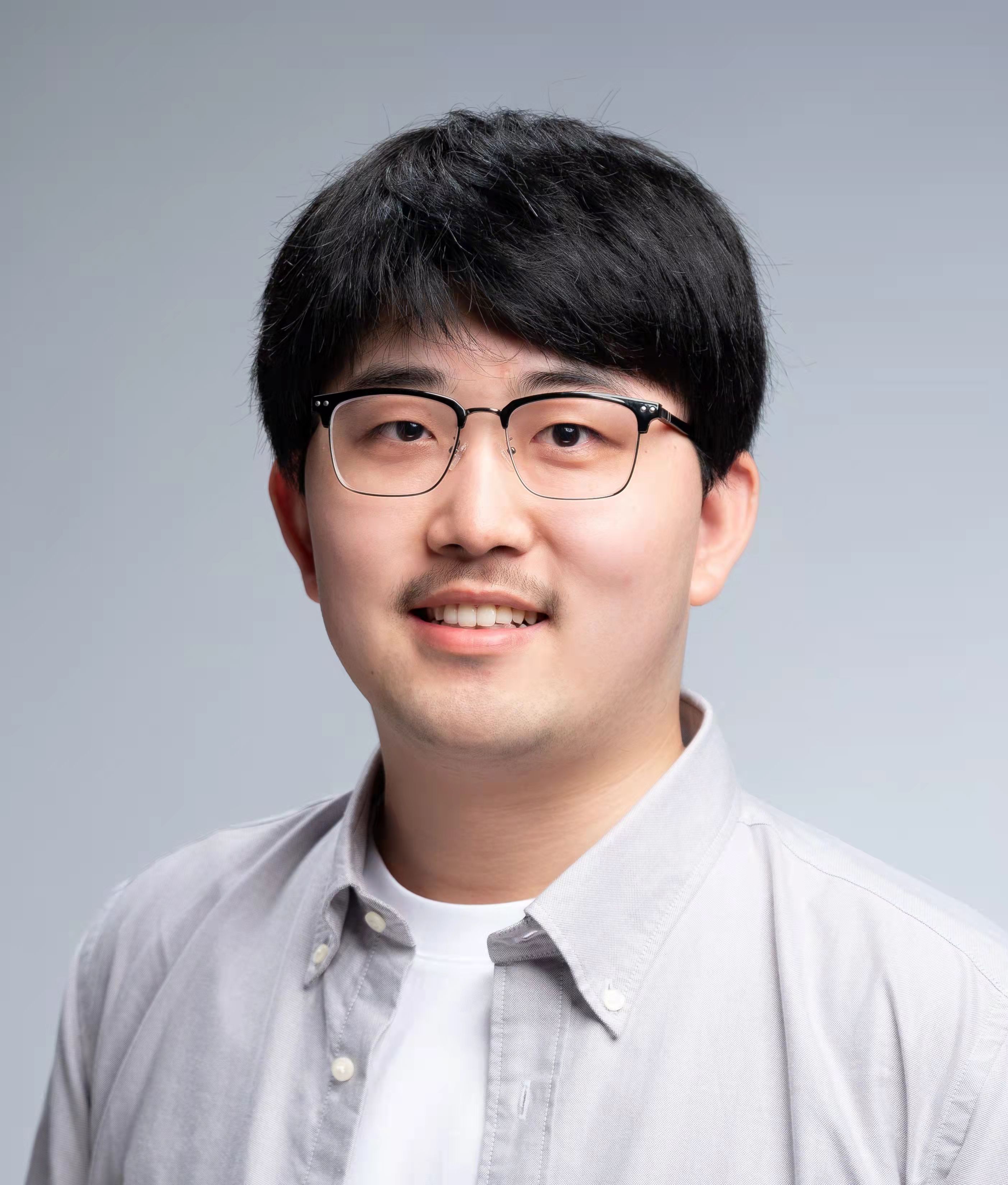}}]{Qiang Sheng} is a Ph.D. student at the Institute of Computing Technology, Chinese Academy of Sciences. He received his B.E. degree from Beijing University of Posts and Telecommunications in 2018. His research interests include fake news detection and fact-checking. He has published over 10 papers in international conferences and journals including ACL, SIGIR, WWW, MM, CIKM, etc.
\end{IEEEbiography}

\begin{IEEEbiography}[{\includegraphics[width=1in,height=1.25in,clip,keepaspectratio]{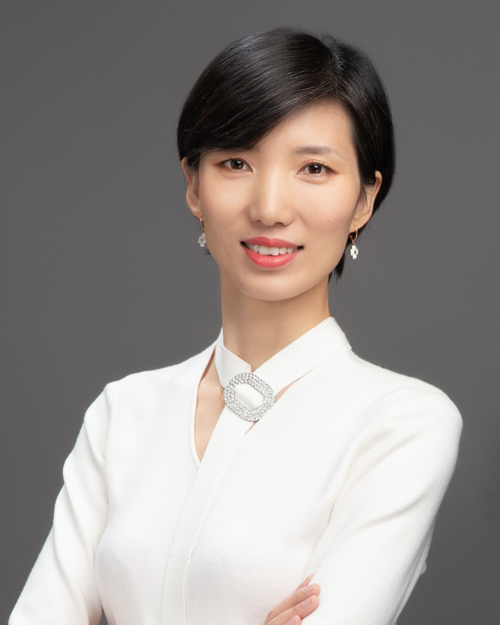}}]{Juan Cao} is a professor in the Institute of Computing Technology, Chinese Academy of Sciences, where she received her Ph.D. degree in 2008. Her research interests include multimedia content analysis and fake multimedia detection. She has over 90 publications in international journals and conferences including TKDE, TIP, KDD, WWW, CVPR, etc. 
\end{IEEEbiography}

\begin{IEEEbiography}[{\includegraphics[width=1in,height=1.25in,clip,keepaspectratio]{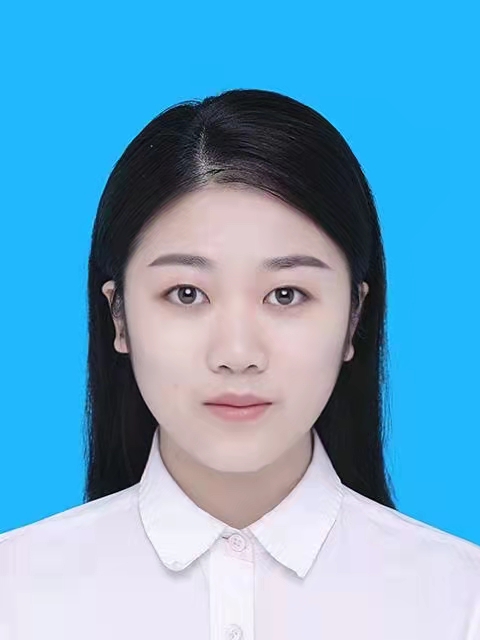}}]{Qiong Nan} is currently pursuing her Ph.D. degree in the Institute of Computing Technology, Chinese Academy of Sciences, Beijing, China. She recieved her B.E. degree from Dalian University of Technology in 2019. Her research interests lie in fake news detection, transfer learning and multi-task learning.

\end{IEEEbiography}

\begin{IEEEbiography}[{\includegraphics[width=1in,height=1.25in,clip,keepaspectratio]{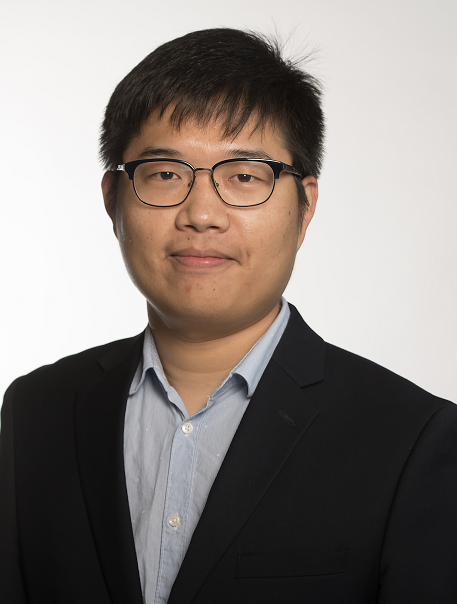}}]{Kai Shu} is a Gladwin Development Chair Assistant Professor in the Department of Computer Science at Illinois Institute of Technology (IIT). He obtained his Ph.D. in Computer Science at Arizona State University in July 2020. He was awarded the ASU Fulton Schools of Engineering Dean’s Doctoral Dissertation Award, CIDSE Doctoral Fellowship 2015 and 2020. His research lies in machine learning, data mining, social computing, and their applications in disinformation, education, healthcare.
\end{IEEEbiography}

\begin{IEEEbiography}[{\includegraphics[width=1in,height=1.25in,clip,keepaspectratio]{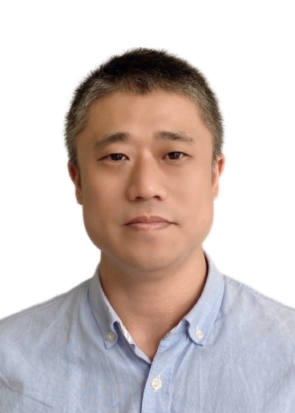}}]{Minghui Wu} received his Ph.D. degree in Computer
Science and Engineering from Zhejiang University.
He is a professor of Computer Science at Zhejiang
University City College. His major interests include
Artificial Intelligence, Big Data, Mobile Application,
and Software Engineering. He has published over 100 papers in journals and international conferences including TKDE, KDD, WWW, etc.
\end{IEEEbiography}

\begin{IEEEbiography}[{\includegraphics[width=1in,height=1.25in,clip,keepaspectratio]{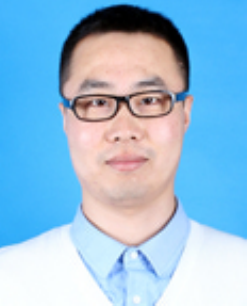}}]{Jindong Wang}
is currently a researcher as Microsoft Research, Beijing, China. He received his Ph.D. degree from Institute of Computing Technology, Chinese Academy of Sciences, Beijing, China. His research interest mainly includes transfer learning, machine learning, data mining, and artificial intelligence. He serves as the reviewers of many journals and conference, including TPAMI, CSUR, and CHI. He is also a member of IEEE, ACM, AAAI, and CCF.
\end{IEEEbiography}

\begin{IEEEbiography}[{\includegraphics[width=1in,height=1.25in,clip,keepaspectratio]{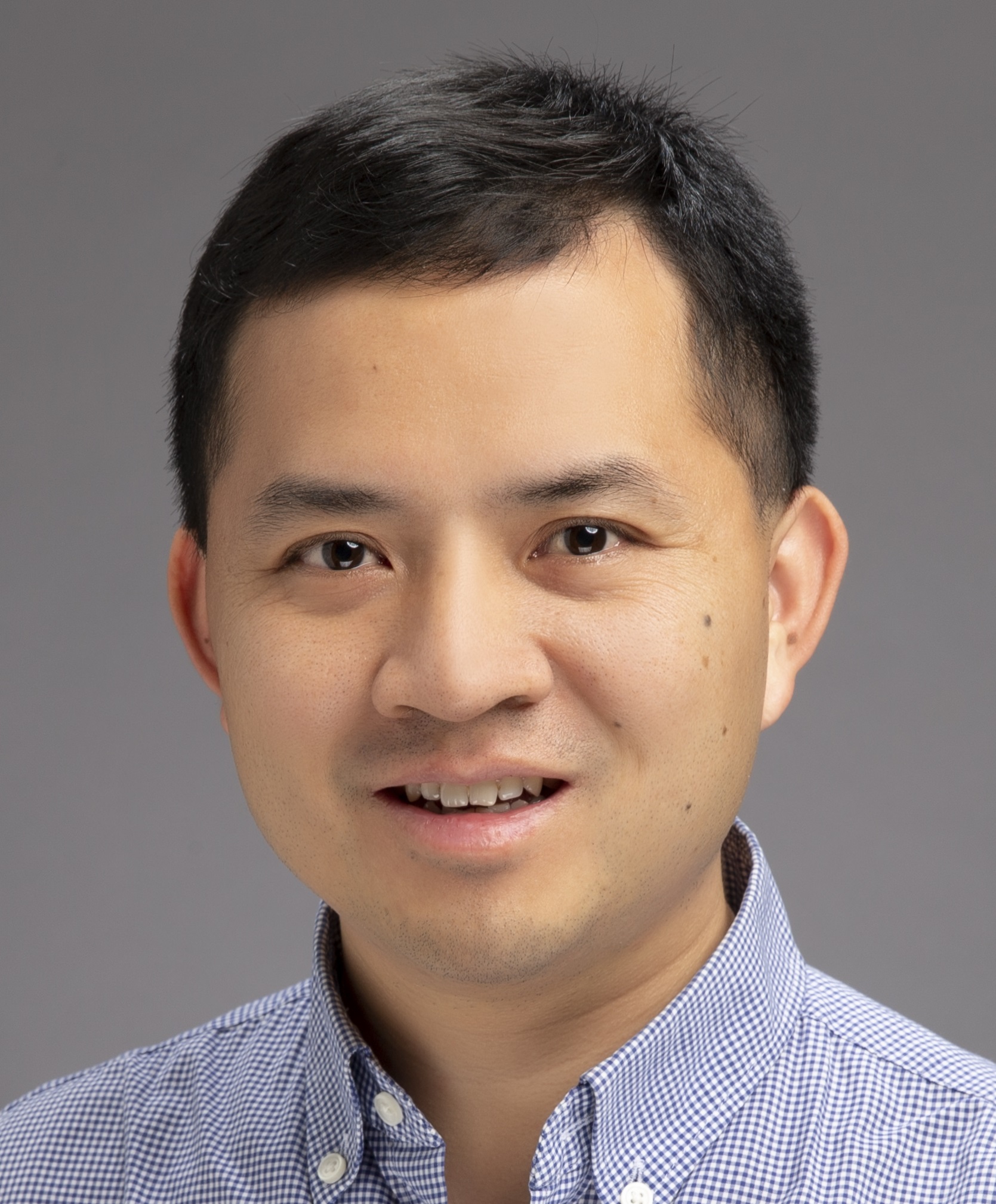}}]{Fuzhen Zhuang}
is a professor in Institute of Artificial Intelligence, Beihang University. His research interests include transfer learning, machine learning, data mining, multi-task learning and recommendation systems. He has published over 100 papers in the prestigious refereed journals and conference proceedings, such as Nature Communications, TKDE, Proc. of IEEE, TNNLS, TIST, KDD, WWW, SIGIR, NeurIPS, AAAI, and ICDE. 
\end{IEEEbiography}




\end{document}